%% file: main.tex
\documentclass[runningheads]{llncs}
\usepackage{hyperref}
\usepackage{subfig}

\usepackage{amsmath}
\usepackage{array}
\usepackage{xcolor}

\usepackage[T1]{fontenc}
\usepackage{graphicx}
\begin{document}
\title{Can LLMs perform structured graph reasoning tasks?}
%
%
\author{Palaash Agrawal\inst{1,2}\orcidID{0000-0001-5902-585X} \and
Shavak Vasania\inst{3}\orcidID{0009-0001-3544-5494} \and
Cheston Tan\inst{1,2}\orcidID{0000-0003-1248-4906}}
\authorrunning{Agrawal et al.}
%
\institute{Center for Frontier AI Research, Agency for Science, Technology and Research, Singapore  \and
Institute of High Performance Computing, Agency for Science, Technology and Research, Singapore \and
UWCSEA Dover High School \\
\email{agrawal\_palaash@cfar.a-star.edu.sg}
}
\maketitle              

\begin{abstract}
Pretrained Large Language Models (LLMs) have demonstrated various reasoning capabilities through language-based prompts alone, particularly in unstructured task settings (i.e. tasks purely based on language semantics). However, LLMs often struggle with structured tasks, because of the inherent incompatibility of input representation. Reducing structured tasks to uni-dimensional language semantics often renders the problem trivial. Keeping the trade-off between LLM compatibility and structure complexity in mind, we design various graph reasoning tasks as a proxy to semi-structured tasks in this paper, in order to test LLMs' ability to use representations beyond plain text. In particularly, we design 10 distinct problems of graph traversal, each representing increasing levels of complexity, and benchmark 5 different instruct-finetuned LLMs (GPT-4, GPT-3.5, Claude-2, Llama-2 and Palm-2) on the aforementioned tasks. Further, we analyse the performance of models across various settings such as varying sizes of graphs, as well as different forms of k-shot prompting. We highlight various limitations, biases and properties of LLMs through this benchmarking process, such as  an inverse relation to the average degrees of freedom of traversal per node in graphs, the overall negative impact of k-shot prompting on graph reasoning tasks, and a positive response bias which prevents LLMs from identifying the absence of a valid solution. Finally, we introduce a new prompting technique specially designed for graph traversal tasks (\textbf{\textit{PathCompare}}), which demonstrates a notable increase in the performance of LLMs in comparison to standard prompting techniques such as Chain-of-Thought (CoT). \footnote{The code for reproducing the results of this paper can be found at  \href{https://github.com/PalaashAgrawal/LLMGraphTraversal}{https://github.com/PalaashAgrawal/LLMGraphTraversal}}

\keywords{Large Language Models \and LLM reasoning \and graph reasoning \and LLM benchmark}
\end{abstract}
\section{Introduction}

\input{content/1.introduction}

\section{Defining Graph Reasoning Tasks}
\input{content/2.graph_definitions}

\section{Results and Analysis}
\input{content/3.results}

\section{Summary}
\input{content/4.summary}

\newpage
\clearpage

\appendix 
\section{Appendix}
\label{appendix}
\input{content/5.appendix}

\end{document}

%% file: content/1.introduction.tex



Large Language Models (LLMs) exhibit interesting reasoning abilities in various types of reasoning tasks \cite{qiao2022reasoning}, including 
arithmetic \cite{wang2022self}, logical \cite{creswell2022faithful}, semantic  \cite{shridhar2022distilling}, and symbolic \cite{saba2023stochastic}. Building on top of just raw reasoning abilities, various techniques, such as chain-of-thought (CoT) and self-consistency~\cite{wei2022chain}, have been proposed to improve the reasoning abilities of LLMs by enhancing the implicit knowledge in the output. However, in relative terms, graph reasoning is not well explored in the domain of LLMs, even though many important subproblems arise in graph reasoning, including subgraph matching \cite{Chen2020can} and shortest path connectivity \cite{goldberg2005computing}. One of the major challenges of grounding graphs through language is rooted in the fact that graphs inherently cover non-linear connectivity between multi-dimensional elements in a condensed format, and hence, reasoning over graphs requires simultaneous tracking of the state of multiple nodes. Broadly in the context of LLMs, graph reasoning is a proxy for understanding relations are established between different entities in a structured setting, and is worth studying, given the impressive emergent abilities observed in recent LLMs. 

The major challenge for LLMs in processing graphs directly arises from the uni-dimensionality of input representations, restricting the scope of reasoning that can be achieved using LLMs \cite{Jin2023Large}. In particular, LLMs are trained on semantic text as next-token predictors, and hence cannot process structured representations \footnote{In this paper, we refer to structured representations as data organized according to a specific schema or format that clearly defines the type, relationship, and arrangement of data elements. This structured format allows for efficient processing using standardized algorithms and tools. In this context, graphs are categorized as structured representations by virtue of node connections. On the other hand, LLMs are designed to process unstructured text representations with an inherent uni-dimensional semantic relationship. By default, LLMs are not equipped to process structured representations through relationship rules.} such as graphs and trees directly, without having to break down the representation into verbally semantic prompts \cite{Liu2023Evaluating}. This process often constrains the selection of tasks, and/or leads to loss of complexity. Hence, efficiently capturing the underlying structure of graphs is particularly challenging for LLMs.

In this paper, we are particularly interested in studying fundamental graph understanding and reasoning abilities of LLMs. While various previous works have studied graph reasoning using LLMs by means of simplified graph structures as enumerated node adjacency lists and connections \cite{Liu2023Evaluating}\cite{Zhang2023LLM4Dyg}\cite{Fatemi2023Talk}, graph tasks such as traversal are inherently reduced to trivial search retrieval problems due to these flattened representations. Hence, \textbf{the ability of LLMs to navigate over more complex and structured representations is still not well explored}. We explore this particular question in-depth in this paper, by going one step further than previous works. Specifically, we prompt LLMs to reason over semi-structured graph representations, instead of plainly verbalized unstructured representations like previous works, while still maintaining the language compatibility of LLMs. \footnote{In this paper's context, semi-structured representations involve breaking down a structured representation into uni-directional text, while still maintaining some form of element relationship. In our case, adjacency matrix representations involve a tabular relation set, which requires tracking both the column and row origin of a node. This structure introduces complexity to the reasoning task. Particularly, we are interested in exploring the ability in LLMs to handle such representational complexity.}



A systematic graph-reasoning benchmarking of pretrained LLMs was performed in \cite{wang2023can}, where various interesting properties were revealed. This included preliminary reasoning capabilities in LLMs in basic graph traversal problems by parsing adjacency lists as input, and the ineffectual nature of advanced prompting techniques in the context of logical deduction. In the aforementioned work, an improvement in LLM reasoning is observed as a result of custom prompting, majorly along the lines of directing the LLM to elaborate on graph connections to facilitate node traversal. However, representation of graphs through adjacency lists does not evaluate the ability of LLMs to navigate through structured representations. Moreover, the design of graph traversal tasks in \cite{wang2023can} is limited in terms of granularity of reasoning evaluation. 

In order to assess the graph reasoning abilities of LLMs in a more comprehensive way, we carry out  graph evaluation through a series of increasingly complex graph problems and settings. Particularly, we construct 10 distinct graph problems requiring multi-hop reasoning and tracking. This includes tree-based graph traversals, grid-based graph-traversals as well as certain special classes of problems. We evaluate five different LLMs (GPT-3.5, GPT-4, Claude-2, Llama-2 and Palm-2). While the OpenAI family of models is thoroughly investigated in various papers, this paper studies the nature of various other contenders. For example, in this paper, we demonstrate that Anthropic's Claude-2 is a good logical reasoning model, only sub-standard to GPT-4 among the various models publically available for access. We also choose to adopt certain design choices that highlight interesting limitations and biases in the training of various models. This includes methods like jumbling the sequence of nodes, as well as testing if the models can identify the absence of a valid solution, or in other words, testing if the models are biased to responding with a non-negative response. 

By carefully examining the logical flow of reasoning adopted by LLMs in order to come to a response, we observe certain limitations that hinder LLMs specifically in the case of graph traversal problems. Using these insights, we propose a novel prompting technique, which we refer to as \textbf{\textit{PathCompare}}, which simplifies the reasoning problem by allowing models to compare different possible solutions. 

In essence, the main contributions of this paper are

\begin{enumerate}
    \item We benchmark five different LLM models on graph reasoning tasks. The graph reasoning tasks are carefully designed in the form of 10 graph traversal problems in increasing orders of complexity. The design of these problems not only assesses the complexity that LLMs can handle, but also highlights interesting properties, including biases introduced due to prompting.
    
    \item To study graph reasoning in depth, we also introduce many variations on top of the graph traversal tasks, including node order (graph size) variation, introduction of weighted and directed connections, and jumbling of node labels. This allows us to guarantee a wide collection of tasks not previously seen by LLMs during pretraining.
    
    
    \item We finally propose a new prompting technique (\textbf{PathCompare}) which improves graph traversal performance across different LLMs. We demonstrate that this prompting technique outperforms standard prompting as well as CoT prompting in the majority (i.e. $>$50\%) of the designed tasks). 
\end{enumerate}

\begin{figure}
    \centering
    \includegraphics[width=\textwidth]{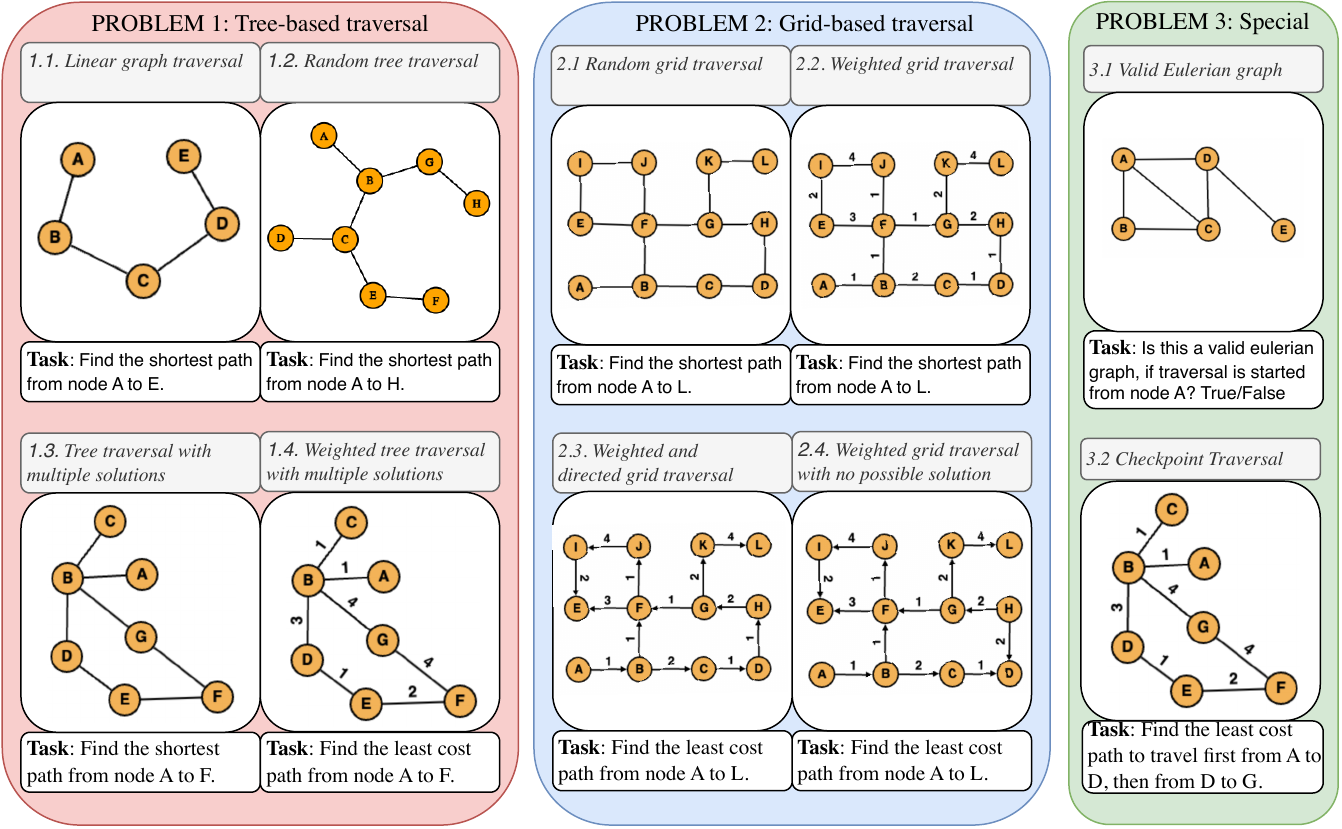}
    \caption{Visualization of all problem categories considered for evaluating LLMs. For each problem, we create 3 variations -- \textit{O(10)}, \textit{O(20)} and \textit{O(20) jumbled}, representing increasing levels of difficulty.}
    \label{fig:graph-visualization}
\end{figure}

%% file: content/2.graph_definitions.tex
A typical graph traversal problem involves navigating a graph $G = \{V,E\}$ between a specified pair of nodes $u,v~\epsilon~V$ through a sequence of edges $e = (e_1, e_2....e_n)~\epsilon~E$. Traversal of a graph $G$ can be formulated in various ways, such as determining connectivity between two sub-graphs, shortest/least-cost path optimization and maximum flow optimization, and so on. However, we identify shortest/least-cost path optimization as one major class of problems involving multi-hop non-trivial reasoning. Analyzing the properties of large language models on such path optimization problems can thus reveal many interesting properties of the models, apart from multi-hop reasoning ability, such as multi-variable state tracking ability, recursive thinking, as well as the ability to reject non-optimal paths. We start by defining the types of graphs that are considered for evaluating traversal properties in LLMs. 

\subsection{Defining graph levels and complexity}

As a starting point, we categorize various types of graphs into two major categories, as described below. 

\paragraph{Problem 1: Tree-based graphs} these graphs are defined as connected undirected graphs, where any two nodes $u,v$ are connected by atmost one edge $e \epsilon \{0, R^+\}$, (where $R^+$ represents positive real number values). In the above representation $e=0$ represents no connection, $e=1$ represents a simple unweighted connection between nodes $u,v$, and all other positive values $e~\epsilon~(R^+ | e\neq 1$) represents a weighted connection between nodes $u,v$. By definitions, no cycles exist in a tree-based graph. However, for the sake of logical grouping of tasks in a generalizable manner, we define \textit{tree-based} graphs as tree-like heirarchical structures with possible exceptions (with respect to cyclic connections). The emphasis is on navigating through these arbitrary node connections to find the shortest or least costly route. Based on this, we formulate the following problems 
\begin{itemize}
        \item \textbf{Problem 1.1: Linear graph traversal} a linear graph is defined as a graph $G(n)$ with $n$ nodes   such that its nodes $V = (v_1, v_2, ... v_n)$ are connected pairwise sequentially ($e_{i,i+1} = 1$), where $e_{i,j}$ represents an edge between nodes $v_i$ and $v_j$. The task involves traversing from node $v_1$ to $v_n$. This is the most trivial form of graph traversal, since there exists only one possible traversal path originating from node $v_1$.
        \item \textbf{Problem 1.2: Random tree traversal}: given a unweighted and undirected tree-based graph $G(n)$ of order $n$, the task involves traversal from node $v_1$ to $v_n$. This involves traversing through and rejecting many dead-end paths. We ensure that there only exists one possible path from node $v_1$ to $v_n$.
        \item \textbf{Problem 1.3: Tree traversal with multiple possible solutions}: given a unweighted and undirected tree-based graph $G(n)$ of order $n$, the task involves finding the shortest path from node $v_1$ to $v_n$, among many possible paths. There exists only one shortest possible path in this tree. 
        \item \textbf{Problem 1.4: Weighted tree traversal with multiple possible solutions} this setting is similar to problem \textit{1.3}, except the edges are weighted. Hence, the task is modified into a least-cost traversal problem. There exists only one possible least cost path.
    \end{itemize}

\paragraph{Grid-based graphs} these graphs are defined as two-dimensional connected graphs $G(M,N)$ of size $M \times N$, where node $u = G(i,j)~\forall (i,j) | (1\leq i \leq M), (1\leq j \leq N) $ is connected to another node $v = G(p,q)$ by atmost one edge $e \epsilon \{0, R^+\}$, (where $R^+$ represents positive real number values, and the definition of edge values are similar to tree-based graphs). A grid-graph is a collection of nodes arranged in a regular grid pattern, with nodes connected to their immediate neighbors in a row-column configuration. By virtue of being two-dimensional,  they are more densely connected than  tree-based graphs, and inherently have cyclic connections (loops) between nodes. Thus, these form a more-challenging class of graph-traversal problems due to the involvement of selection among a higher number of possible solutions. Based on this, we formulate the following problems. 

\begin{itemize}
        \item \textbf{Problem 2.1: Random grid traversal} given a unweighted and undirected graph $G(M,N)$ with dimensions $(M,N)$, the problem involves finding the shortest path of traversal from node $G(1,1)$ to node $G(m,n)$. There exists only one shortest possible path per grid. 
        \item \textbf{Problem 2.2: Weighted grid traversal} This problem is similar in setting to problem \textit{2.1} except the edges are weighted. Hence the task evolves into a least-cost path traversal from node $G(1,1)$ to $G(m,n)$. There exists only one least-cost path per grid. 
        \item \textbf{Problem 2.3: Directed and weighted grid traversal} This problem is similar in setting to problem \textit{2.1} except all edges are weighted as well as directed. Hence the problem evolves into finding a valid least-cost traversal from node $G(1,1)$ to $G(m,n)$. 
        \item \textbf{Problem 2.4: Directed grid traversal with no possible solution} This problem involves a directed grid $G(M,N)$, such that no valid path exists because of randomized directions. The goal of the problem is to evaluate whether LLMs are capable of evaluating directional conflicts, and come to the conclusion that no path is valid.
    \end{itemize}

By incrementally adding constraints to both the problem categories above, we create an analysis framework to determine the graph reasoning abilities of LLMs in increasing levels of complexity. Between the two problem categories, problem 2 poses a more challenging problem than problem 1, due to a larger number of average degrees of freedom of traversal per node.

\paragraph{Problem 3: Special problems} Apart from the two aforementioned categories of problems, we define two special tree-based traversal problems.
\begin{itemize}
    \item \textbf{Euler walk}: Given a valid Euler graph $G$, an Euler path is defined as a path which covers all edges $E$ of the graph exactly once. The Eulerian path traversal is a more viable alternative to the \textit{Hamiltonian path} traversal, which involves travelling all nodes $U$ of the graph exactly once, since the former can be evaluated computationally by a finite set of conditions, unlike the latter, which is an NP-complete problem and cannot be evaluated using a polynomial time algorithm. Hence, we formulate the problem as follows. 
    \begin{itemize}
        \item \textbf{Problem 3.1: Valid Euler graph} Given a tree-based graph $G(n)$ with $n$ nodes, the task involves identifying whether or not a valid Euler path is possible, given a starting node $v_i$. 
        
    \end{itemize}
    \item \textbf{Checkpoint traversal} This problem involves traversal of a tree-based graph $G$ between two nodes $u$ and $v$, such that a specific node $w$ must be part of the traversal path. Mathematically, this is equivalent to breaking down the traversal into two distinct problems -- traversal from $u$ and $w$, and traversal from node $w$ and $v$. The goal of this cascaded traversal problem is to evaluate whether a single LLM response can simultaneously solve multiple problems. We hence formulate the following problem.

    \begin{itemize}
        \item \textbf{Problem 3.2: Cascaded graph traversal} Given a weighted tree graph $G(n)$ with n nodes, find the least cost path to traverse from node $v_1$ to a randomized node $v_i$, and then from $v_i$ to $v_n$.
    \end{itemize}
\end{itemize}

\subsection{Graph generation, prompting and evaluation}
\paragraph{Automated graph generation}
Graphs and their corresponding solutions are automatically generated as adjacency matrices with alphabetically labeled nodes. We deliberately avoid adjacency lists (a more standard practice of computational representation), to avoid triviality, since a direct listing of individual node connections simplifies the complexity for the LLM, thereby hindering the ability to assess its raw reasoning capabilities. We ensure that  there exists only one unique solution for all problems, so that evaluation can be unambiguous. For problem \textit{3.1} (Euler walk), we convert it into a True/False problem due to the lack of a guaranteed single Eulerian path, ensuring a balanced distribution of valid and invalid cases. All remaining problems require LLMs to provide a sequence of nodes in response. Thus, while lower LLM accuracies are expected, a more comprehensive evaluation can be performed. Across all problems, we generate three variations (or orders) of problems, depicting increasing levels of complexities.
\begin{enumerate}
    \item \textit{O(10)}: This refers to graphs that have a total number of nodes varying around $n=10$, or between 5 and 15 (inclusive) to be exact. Hence the problem is formulated as traversal between \textit{A} and \textit{J} when $n=10$, and so on, in the case of problem categories 1 and 2.

    \item \textit{O(20)}: This refers to graphs which have a total number of nodes varying around $n=20$, or between 16 and 26 (inclusive). Hence, when $n=20$, the problem is formulated as traversal between node \textit{A} and \textit{T}, and so on, in the case of problem categories 1 and 2. 

    \item  \textit{O(20) jumbled}: In contrast to the previous cases, where node labels follow a near-ordered sequence in depth-first traversal, the labels of the nodes in 
\textit{O(20)} are randomized to increase complexity. Although the adjacency matrix remains alphabetically ordered, the solution sequence is entirely shuffled. This approach allows us to assess whether LLMs depend on training biases, such as node order, when generating solutions.
\end{enumerate}

Further, we ignore evaluating problem 3.1 (valid euler graph) for O(20) and O(20) jumbled graphs. This is because generating larger graphs with guaranteed eulerian conditions are computationally highly expensive. 

\paragraph{Prompting techniques} In order to maintain structure of graphs without explicit flattening and enumeration of nodes/connections, we represent graph structures as \textbf{adjacency matrices}, rather than simpler adjacency lists. This kind of structure is not only more compact, but tests the ability of LLMs to perform multi-hop reasoning, and tracking value states more rigorously. A detailed description of our base prompting method is described in the Appendix. 

For prompting, our goal was to ensure that all models are given the same prompts for fair evaluation. For all models, we test graph reasoning abilities across 3 k-shot settings -- \textit{zero-shot} setting ($k=0$), \textit{one-shot} setting ($k=1$) and \textit{three-shot} setting ($k=3$) \cite{brown2020language}. To evaluate and compare the preliminary graph reasoning abilities of LLMs, we avoid any specialized prompting techniques (such as self-consistency \cite{wang2022self} and other related techniques). However, later in the paper, chain-of-thought prompting \cite{wei2022chain} is introduced, majorly to compare the impact specialized prompting on graph reasoning. 


\paragraph{Partial Credit evaluation} Apart from an automated evaluation, we go one step further and manually evaluate problem categories 1 and 2 (except 2.4, where partial credit does not apply) for partial credit. We define partial credit as the fraction of nodes that the evaluator got correct out of all the nodes in the ground truth solution before predicting to a wrong node. This is only applicable in cases where the model does not get a full score of 1.0 from the primary evaluator. This metric allows us to evaluate the reasoning ability and problem solving technique of LLMs with greater granularity.

\subsection{Selection of LLMs}
In order to include a broad range of models, we select five models from four families of models (in terms of architecture and training data/method). This includes -- OpenAI's \textbf{GPT-3.5} and \textbf{GPT-4}  \cite{openai2023gpt4}, Anthropic's \textbf{Claude-2} \footnote{It should be noted that Claude-3 was not released at the time of conducting these experiments.}\cite{wu2023comparative} , Google's \textbf{Palm-2} \cite{anil2023palm}, and Meta's \textbf{Llama-2} \cite{touvron2023llama} with 13B parameters. Since LLama-2 is a raw foundational model, we use an instruction-fine tuned version (Hermes) released by \textit{Nous Research}.

%% file: content/3.results.tex
Using the prompting and evaluation technique defined above, we evaluate models over the defined problems. We evaluate the results of graph reasoning for all problems by calculating
averaged and normalized binary accuracies across 10 examples for each setting, and go from 0 to 1 in increments of 0.1. A model response is only marked correct if it predicted the shortest/least-cost traversal path correctly and completely. Along with binary accuracies, partial accuracies are also calculated. While we use binary accuracy to filter out models with weak reasoning abilities, we use partial accuracy for more granular comparison. These results can be visualized in \textbf{Fig. \ref{fig:o10-0shot}}. A detailed tabulation of these results are included in the \textbf{Appendix tables \ref{tab:prob1}} through \textbf{\ref{tab:prob3}}. 

\begin{figure}
    \centering
    \includegraphics[width=0.6\linewidth]{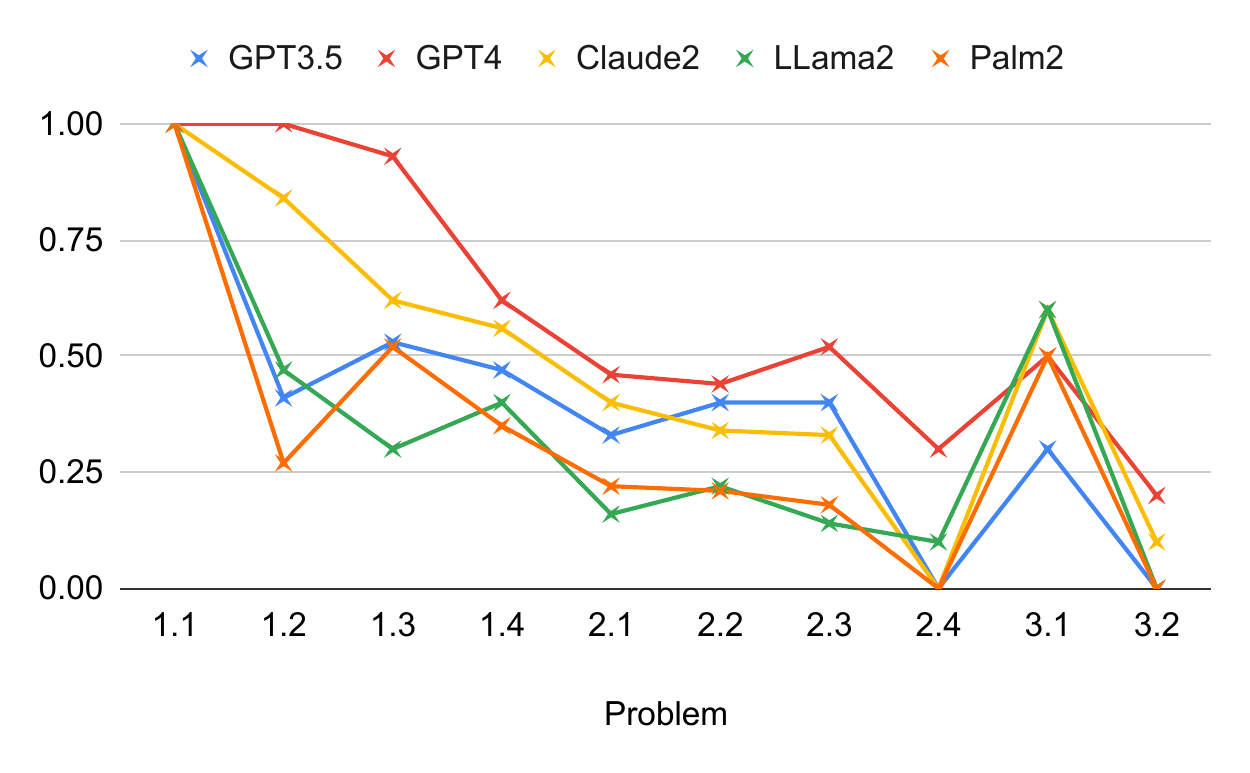}
    \caption{Baseline comparison of all model families on O(10) graph problems in 0-shot settings using partial accuracy. Partial accuracy gives a more granular insight into the performance of models, versus binary accuracy.}
    \label{fig:o10-0shot}
\end{figure}

\begin{figure}[!ht]
    \centering
    \subfloat{{\includegraphics[width=0.45\textwidth]{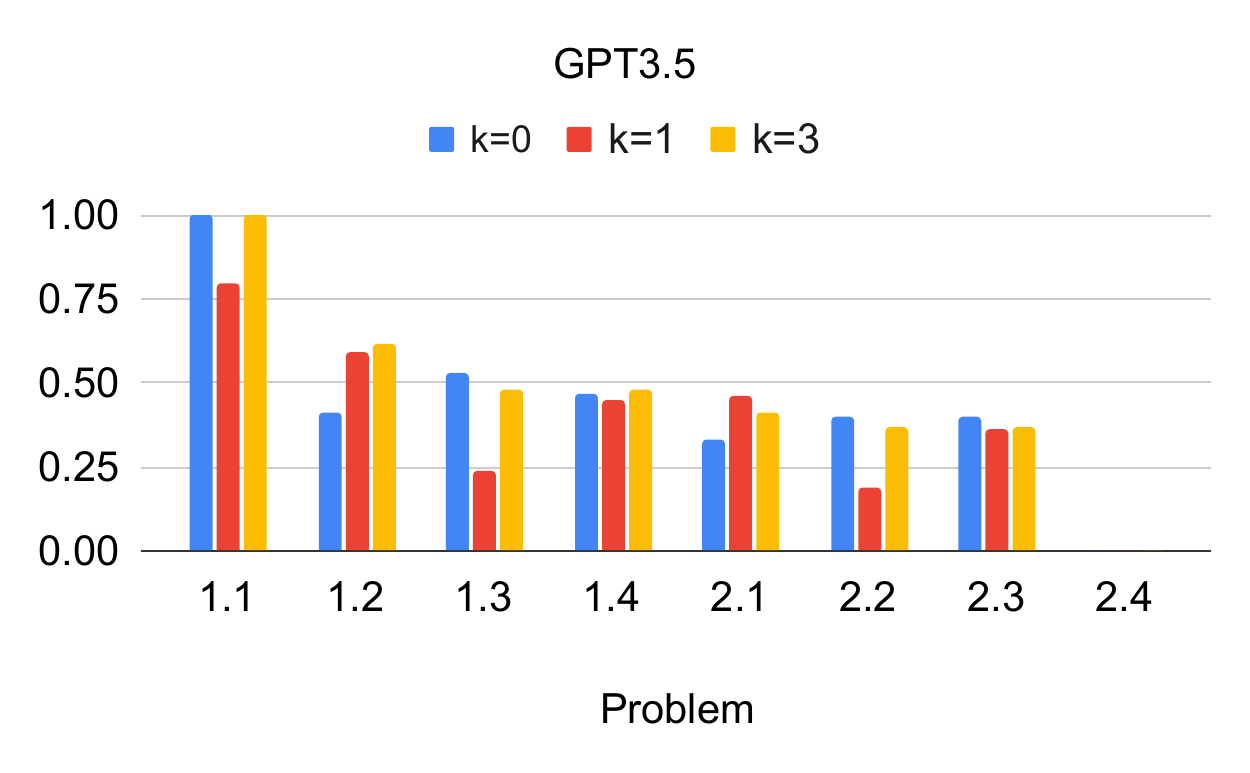} }}%
    \qquad
    \subfloat{{\includegraphics[width=0.45\textwidth]{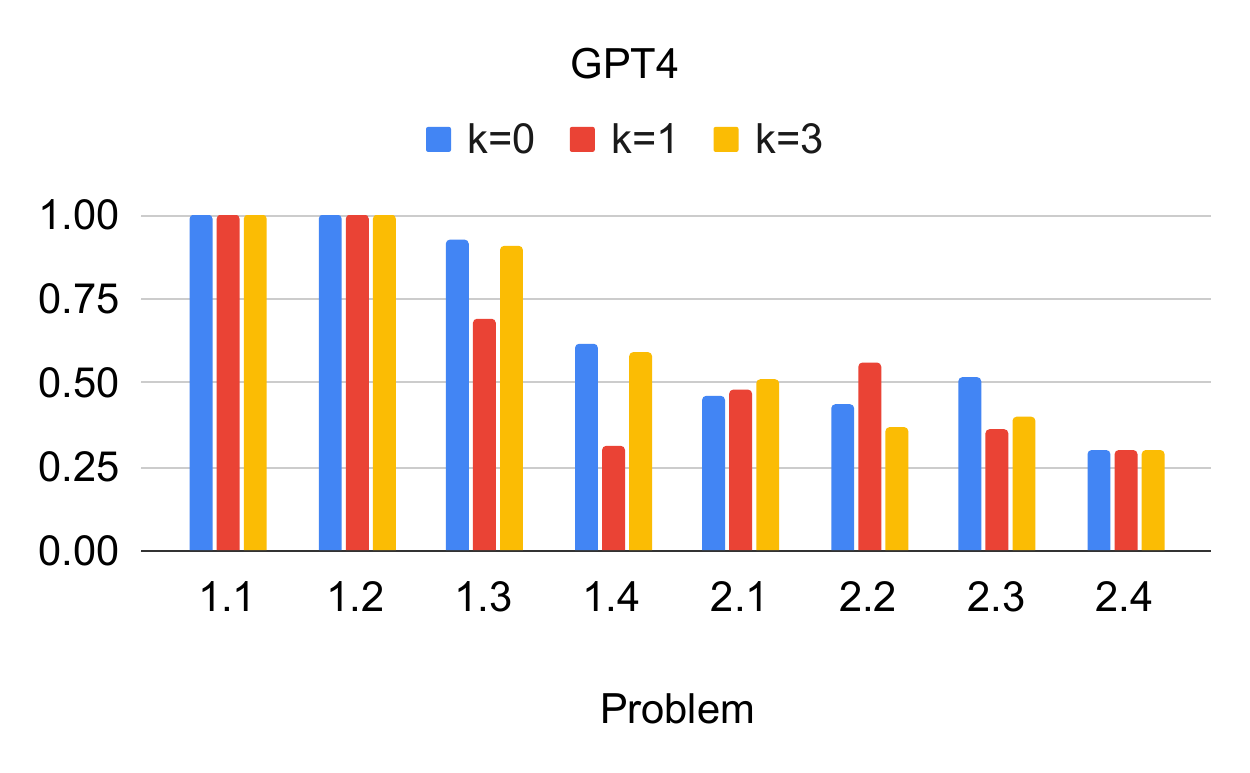} }}%
    \qquad
    \subfloat{{\includegraphics[width=0.45\textwidth]{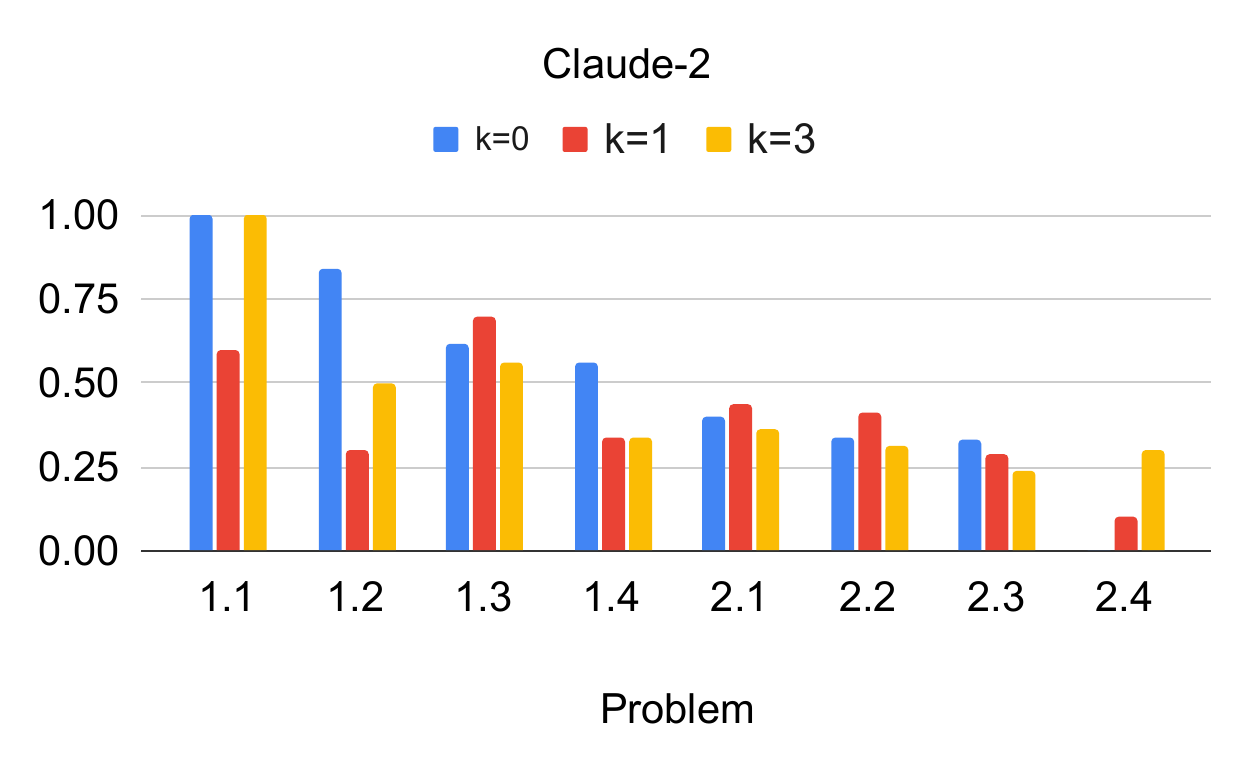} }}%
    \qquad
    \subfloat{{\includegraphics[width=0.45\textwidth]{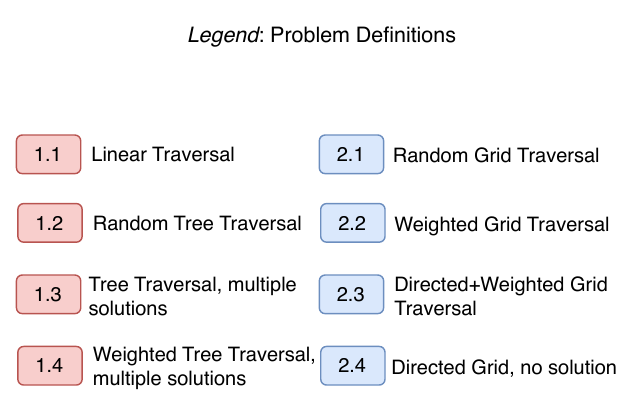} }}%
    
    \caption{Comparison of the effect of k-shot prompting on various LLMs. We observe that in more than half of all tasks, few-shot prompting leads to either a drop or an insignificant ($\leq$ 5\%) improvement in accuracy in comparison to 0-shot prompting (depicted in blue). This is observed in 6/8 tasks for GPT3.5, 6/8 tasks for GPT4 and 5/8 tasks for Claude-2.}%
    \label{fig:k-shot comparison}%
\end{figure}

\begin{figure}[!ht]
    \centering
    \subfloat{{\includegraphics[width=0.45\textwidth]{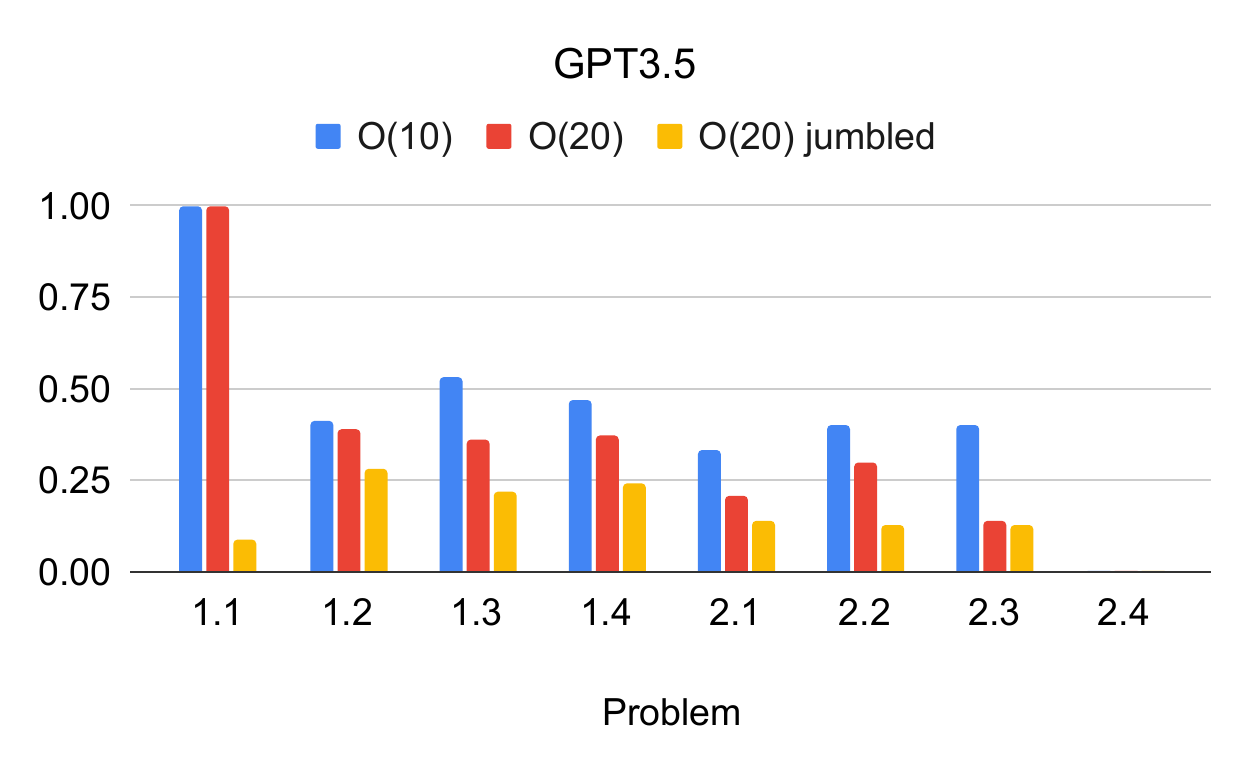} }}%
    \qquad
    \subfloat{{\includegraphics[width=0.45\textwidth]{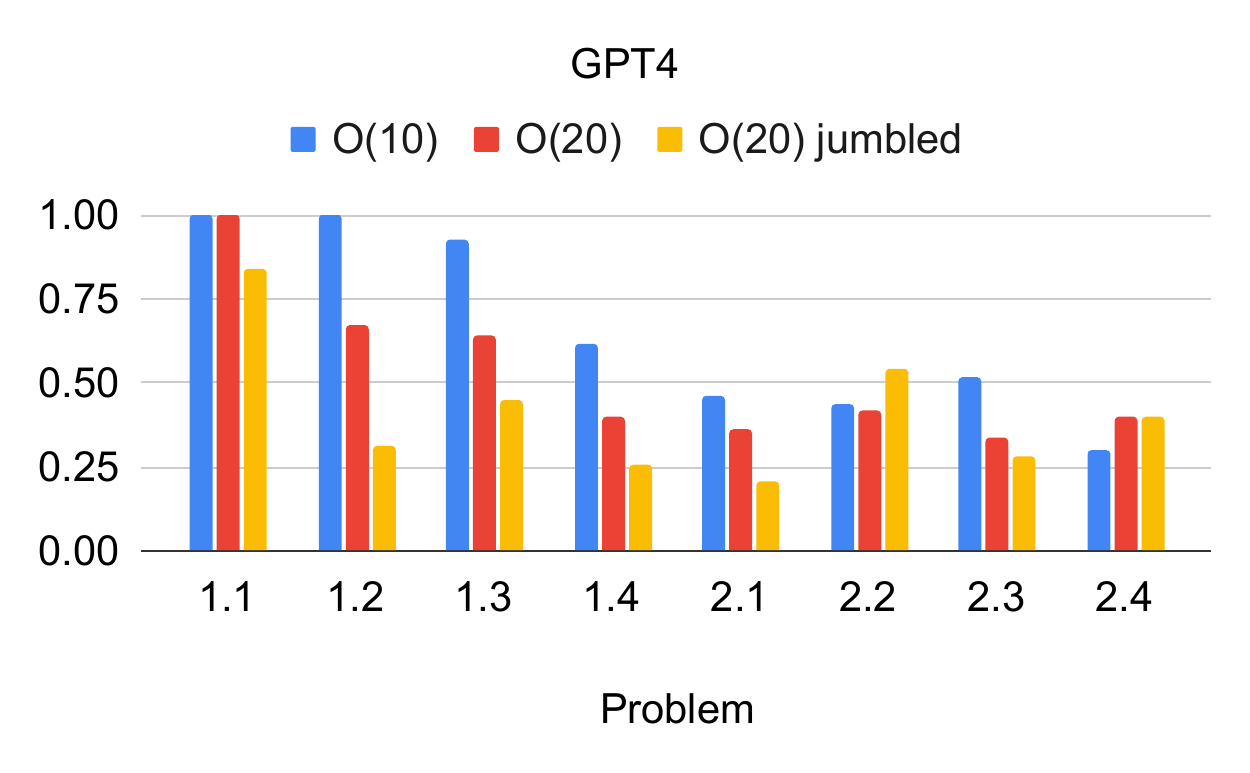} }}%
    \qquad
    \subfloat{{\includegraphics[width=0.45\textwidth]{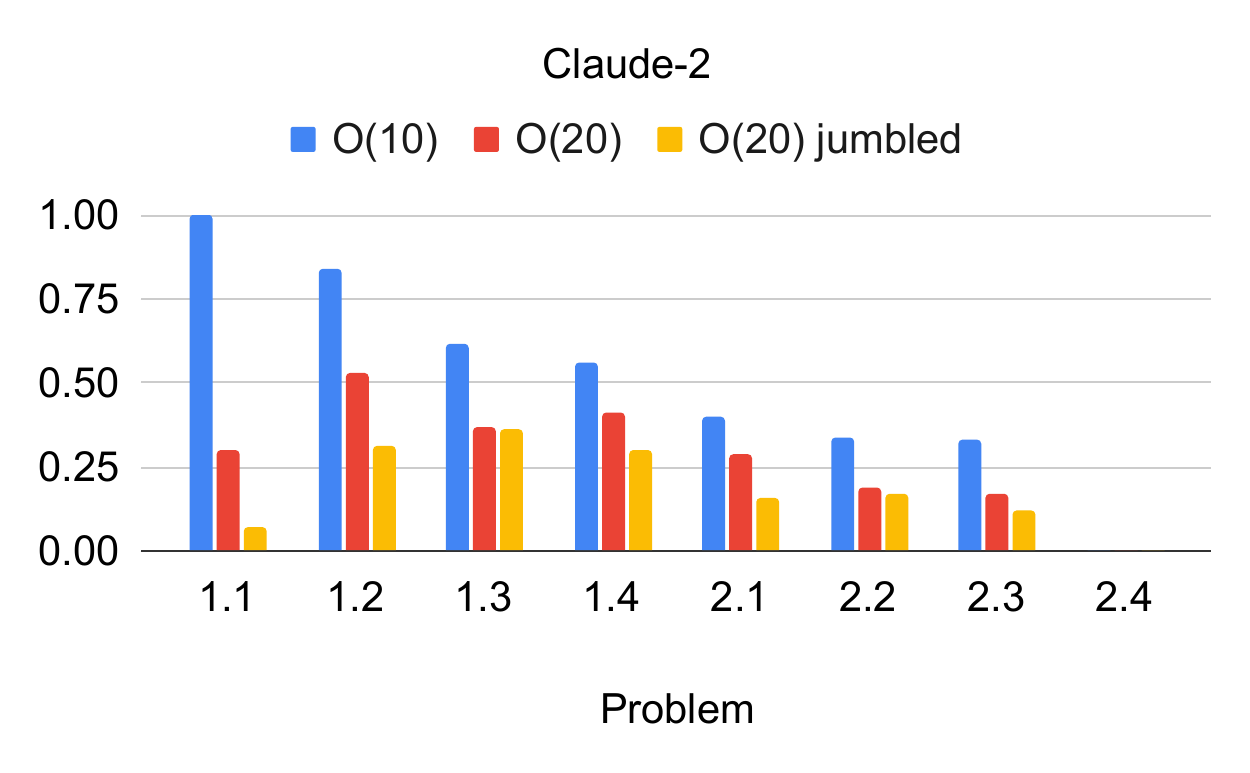} }}%
    \qquad
    \subfloat{{\includegraphics[width=0.45\textwidth]{figures/legend.pdf} }}%
    
    \caption{Comparison of graph order performance in various LLMs. LLM performance accuracy consistently drops as the order of the graph is increased. Also, while keeping the order of graph magnitude at 20, a drop is observed in performance when the nodes are jumbled, depicting a bias in models to expect node order to be alphabetically arranged.}%
    \label{fig:order comparison}%
\end{figure}

\subsection{Preliminary technical limitations}
Before analysing the performance of various LLMs on different LLMs, we observe some overall characteristics of models. Particularly, we observe issues with Llama-2 and Palm-2. In the case of Llama-2, 3-shot setting for O(20) graphs could not be evaluated, because of the limited context-window length of the model (4,096 tokens). While this issue was also seen with the default GPT-3.5 model, a GPT-3.5 variant with a larger context window (16,383 tokens) was able to solve the issue. In the case of Palm-2, we observe that beyond a certain length of input (in the case of 1-shot and 3-shot prompts), the model tends to not respond with any solution, but with a generic statement related to the model's inability to solve the problem.

\subsection{Overall observations from Problem 1 and Problem 2}
Through a first, glance, we can easily point out that all models generally perform better on tree-based graphs (problem 1) than grid-based graphs (problem 2). Further, we notice a general drop in performance in random tree traversal (problem 1.2) across O(10) and O(20) graphs. This clearly indicates that a greater number of average degrees of freedom for traversal per node has an inverse correlation with reasoning capability.

\paragraph{Model specific observations}
Although all models perform decently on trivial graph traversal tasks (e.g. problem 1.1), differentiation among models is quickly observed as more constraints are added. Referring to \textbf{Fig. \ref{fig:o10-0shot}}, a general aggregate trend is observed where some reasoning abilities are observed in the case of GPT4, Claude-2, and GPT-3.5 (in this order of decreasing relative reasoning abilities). On the other hand, Llama-2 and Palm2 perform under 30\% on the majority ($\geq$ 50\% of the settings) of the tasks, depicting poor reasoning abilities. In more complex settings (such as O(20) graphs and O(20) jumbled graphs, not presented in this paper for simplicity and readability, but present in the code repository), an almost random-like performance is seen in these models. An anomaly is observed in problem 3.1 where a spike is observed in the performance of the aforementioned 2 models. However, since this particular problem is a True/False problem with a random baseline accuracy of 50\%, the spike is rendered meaningless. Keeping this in mind, in the remainder of the paper, we primarily focus on GPT4, Claude-2, and GPT-3.5, since these contenders depict fairly reasonable graph reasoning abilities. However, all the results are openly available for analysis in our code repository.




\subsection{Weighted vs unweighted graph traversal}
Adding weights to graph edges in least-cost traversal problems introduces significant complexity, requiring models to track cumulative weights. Unsurprisingly, we observe a general trend of dropping performance as soon as weights are added to trees (problems 1.3 vs 1.4) and well as grids (problems 2.1 vs 2.2). GPT-4 and Claude-2 demonstrate some ability to solve weighted traversal, demonstrating their innate ability to track numerical variable states in parallel to graph paths.
On the other hand, chain-of-thought prompting \cite{wei2022chain} in GPT-3.5 demonstrates that the models tend to neglect weights and solve problems using an unweighted approach. The remainder of the models such as Llama-2 and PaLM 2 tend to respond with a description of the algorithm underlying the solution, rather than the solution itself. This is especially observed in 0-shot settings.

\subsection{Effect of k-shot prompting}
In \textbf{Fig. \ref{fig:k-shot comparison}}, we compare the effect of k-shot prompting across 8 tasks, and observe that in the majority of the tasks, k-shot prompting has either no statistically significant effect, or a negative effect on reasoning accuracy. In general, while few-shot prompting is helpful in response format shaping, it is of no particular aid for analytical tasks such as graph reasoning. Carefully examining individual responses, we observe models (even the more powerful ones like GPT4) confusion between different examples, and responding to k-shot examples rather than the relevant graph traversal question.  


\subsection{Effect of increased graph order and jumbling node order}
In \textbf{Fig. \ref{fig:order comparison}},We analyze the impact of different graph orders and complexities, specifically comparing O(10), O(20), and O(20) jumbled graph settings. We consistently observe across all models that a performance drop is observed when graph order is increased. An interesting insight enlightened when manually analyzing individual solutions is that poor performance in O(20) graphs in comparison to O(10) graphs is not solely due to increased computational demands but also because models often omit higher node labels (e.g., J, K, L, M and so on) when labels are alphabetically ordered, indicating training on lower-order graphs.

This behavior is amplified in the case of jumbled node sequences, where even trivial linear traversal becomes challenging, highlighting a strong model bias towards ordered node sequences, even in few-shot settings. GPT-4, while occasionally applying valid logical principles, still struggles across grid-based traversals.

\subsection{Positive response bias within LLMs}
Through a special problem (problem 2.4), we observe that all models also consistently fail to recognize when no solution is possible. An interesting observation is that models fail to recognize when no valid solution is possible even in few-shot prompt settings, considering it is expected that this method should induce a formatting bias in the response. This indicates a clear training regime flaw among models to prefer avoiding empty responses at the cost of predicting wrong solutions. However, given that GPT-4 and Claude-2 do possess some valid logical reasoning and state tracking abilities, we observe some ability to predict the inexistence of any valid solution in these models. Surprisingly, in the case of few-shot prompt (where we prompt models with examples of \textbf{valid} graphs with solutions, the aforementioned models tend to identify the absence of a valid path with better accuracy (see \textbf{Fig. \ref{fig:k-shot comparison}}). 

\subsection{Performance of LLMs over special problems}
As mentioned earlier, in this paper, the eulerian problem (problem 3.1) is a binary problem, and thus a random classifier is expected to perform with an accuracy of 0.5. We observe slightly better performance than random only in the case of Claude-2 and Llama-2. \textbf{However, in all other models, the performance over identification of a valid eulerian graph is equal to or less than random.
} Analysis of the responses manually highlights that models tend to give a positive response. 




\section{PathCompare prompting}
As clearly seen in the previous sections, LLMs struggle with graph reasoning. However, this is expected, since Large Language Models are, in essence, token predictors, and do not inherently possess the ability to go back to the prompt recursively to self-correct. By prompting LLMs to directly generate the optimal path to a graph traversal problem, we inherently prevent the model from a systematic analysis of the graph traversal problems. 

A more advanced prompting technique such as Chain-of-Thought reasoning \cite{wei2022chain} has more or less no effect on the performance of LLMs, specifically on graph reasoning. We observe, that even without a prompt augmentation that compels the model to list down its reasoning, LLMs come to a conclusion through an algorithmic walk (such as Dijkstra's algorithm, or breadth-first-search). Essentially, it is not the lack of the flow of reasoning that prevents LLMs from coming to a consistent and accurate traversal solution, but the lack of ability to explore and compare with other valid solutions. 

\begin{figure*}[!ht]
    \centering
    \subfloat{{\includegraphics[width=0.45\textwidth]{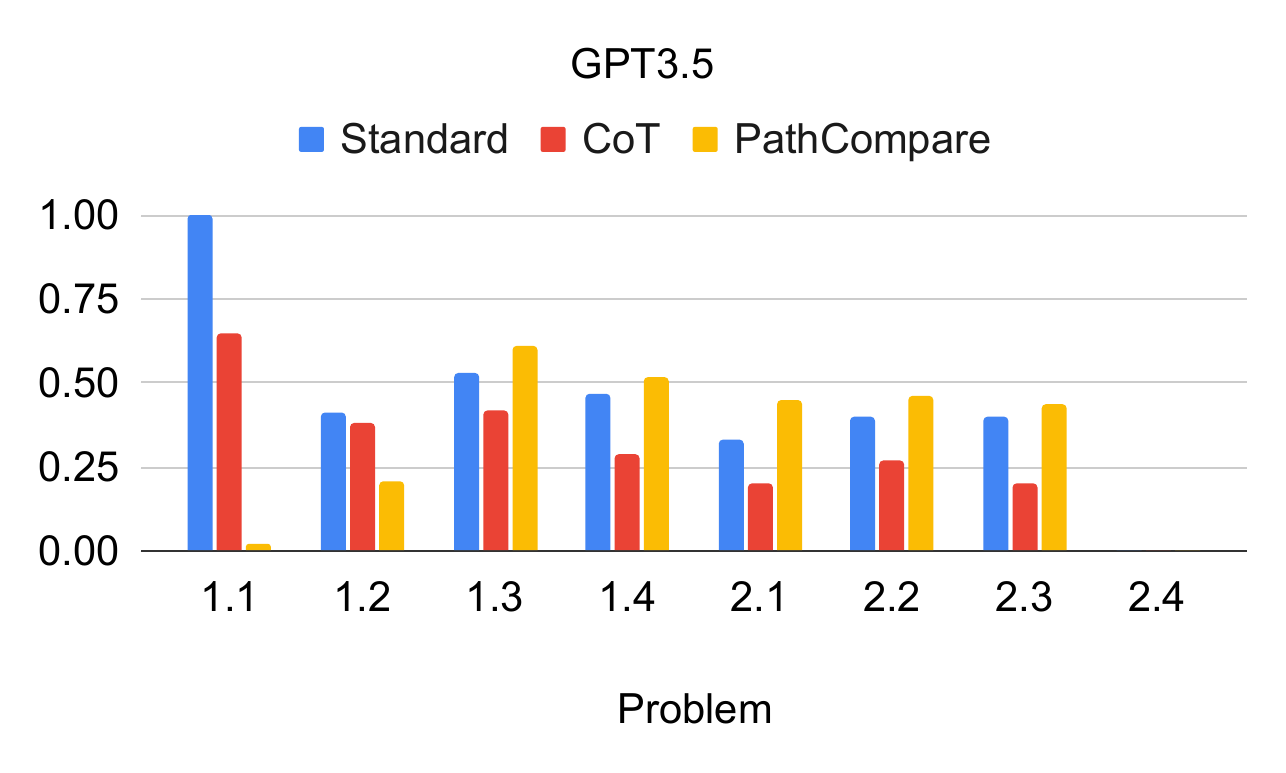} }}%
    \qquad
    \subfloat{{\includegraphics[width=0.45\textwidth]{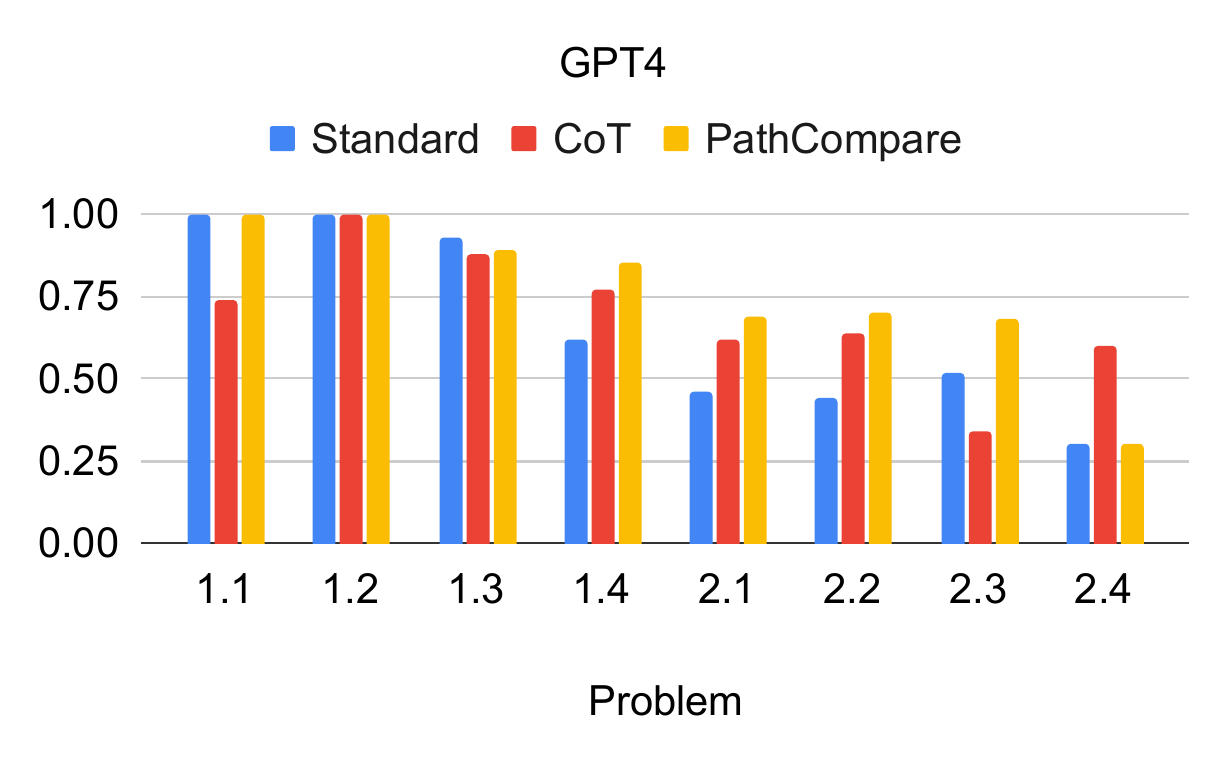} }}%
    \qquad
    \subfloat{{\includegraphics[width=0.45\textwidth]{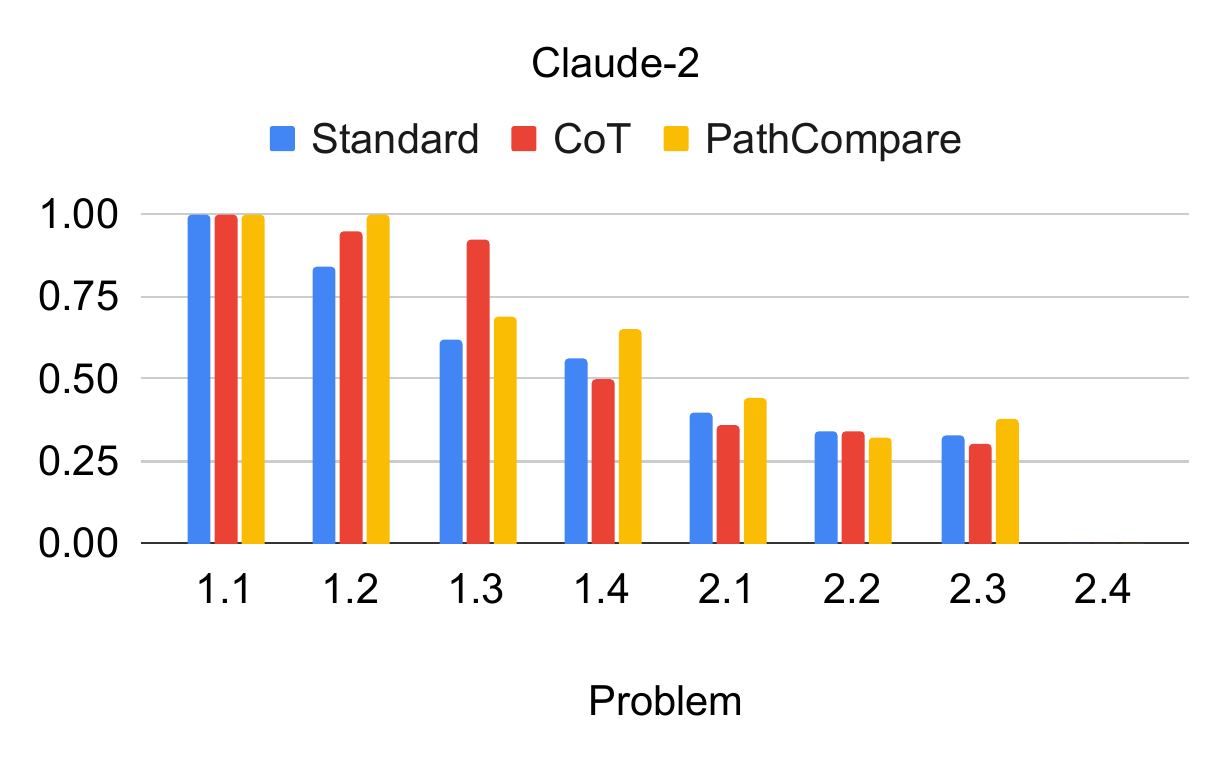} }}%
    \qquad
    \subfloat{{\includegraphics[width=0.45\textwidth]{figures/legend.pdf} }}%

    \caption{Comparison of different prompting techniques. Our proposed prompting technique (\textit{PathCompare}) demonstrates an improvement of accuracy in the majority of tasks across all models, i.e. 5/8 tasks for GPT3.5, 6/8 tasks for GPT4 and 5/8 tasks for Claude-2. However, one limitation of PathCompare is that it enhances positive response bias, as observed in the case of problem 2.4.}%
    \label{fig:pathcompare}%
\end{figure*}

Keeping these requirements in mind, we propose a novel prompting technique specifically for graph reasoning (traversal) tasks, which we refer to as the \textbf{\textit{PathCompare}} technique. By simply adding the line \textit{``Let's list down all the possible paths from node \{X\} to node \{Y\}, and compare the cost to get the answer."}, we observe a significant improvement in the graph reasoning performance of all LLMs (see \textbf{Fig. \ref{fig:pathcompare}} for visualization and \textbf{Appendix Table \ref{tab:pathcompare}} for detailed analysis), in comparison to standard prompting and other forms of prompting such as CoT (see \textbf{Appendix Table \ref{tab:CoT}}). Through this prompt, we shape an LLM response to first list down multiple paths between two specified nodes. Once multiple paths are enumerated, the task boils down to a retrieval-cum-comparison task, which increases the probability of a more accurate response. 

%% file: content/4.summary.tex
Through this paper, we probe deeper into the graph-reasoning capabilities of various LLMs. This study adopts a more granular approach towards analysing LLMs, in comparison to previous works \cite{wang2023can}. First, we select a wider range of models from different model families (i.e. different architectures, training data and methods and distinct fine-tuning methods), and systematically compare the different models over reasoning tasks. Our problem design evaluates these models over increasingly complex problems and added constraints. Through this process, we highlight some interesting properties of LLMs. First off, we highlight various type of constraints that lead to performance drop (such as weighted edges and unordered node sequences). We also conclude that k-shot prompts are unhelpful for analytical tasks such as graph traversal. We also highlight how various models are biased towards providing a positive response. This, hence, leads to these LLMs providing wrong responses, even when no valid solution exists. Secondly, we highlight the root of poor graph-based reasoning in LLMs, and link this to the inability to backtrack solutions to compare and come to an optimal conclusion. Building on top of this, our novel prompting technique \textit{PathCompare} helps mitigate this issue and improve graph reasoning abilities in LLMs.

%% file: content/5.appendix.tex
\subsection{Prompting}

\subsubsection{Base Prompt}
We prompt models using adjacency matrices, labeled sequentially as \textit{A,B,C...}, with 0 indicating no connections between 2 nodes, and 1 indicating otherwise. The model is asked to find the shortest path node A, to the last sequential node. An example prompt is as follows. 

\textit{ Given is the adjacency matrix for a unweighted undirected graph containing 10 nodes labelled A to J. The value corresponding to each row M and column N represents whether there is a connection between the two nodes, where 0 means no connection. }  

\textit{What is the shortest path from node A to node J? Return the sequence of nodes in response.}




\begin{center}
     \begin{tabular}{ccccccccccc}
& ~A& ~B & ~C &~D &~E &~F& ~G &~H &~I &~J\\ 
~A & ~0& ~1& ~0& ~0& ~0& ~0& ~0& ~0& ~0& ~0 \\
~B& ~1& ~0& ~1& ~0 &~0 &~0 &~0 &~0 &~0 &~0 \\

~C& ~0& ~1& ~0& ~1& ~0& ~0& ~0& ~0& ~0& ~0 \\
~D& ~0& ~0& ~1& ~0& ~1& ~0 &~0 &~0& ~0& ~0 \\
 ~E &~0 &~0 &~0 & ~1& ~0& ~1 &~0& ~0& ~0& ~0 \\
 ~F& ~0 &~0 &~0 &~0 &~1& ~0& ~1& ~0& ~0 &~0 \\
 ~G &~0 &~0& ~0& ~0& ~0& ~1& ~0& ~1& ~0& ~0 \\
 ~H &~0& ~0& ~0& ~0& ~0& ~0& ~1& ~0 &~1& ~0  \\
 ~I &~0& ~0& ~0& ~0& ~0& ~0& ~0& ~1& ~0& ~1 \\
 ~J& ~0& ~0& ~0 &~0 &~0 &~0& ~0& ~0 &~1 &~0 \\
\end{tabular}
\end{center}

Evaluation of the output of the models was performed manually, in order to factor in the variation of the output format by different models. In our code, input/output pairs (prompt/solution pairs) are automatically generated, which allows us to create a large scale of variations. An example of the output string generated for this corresponding problem is 
\begin{center}
    \textit{A -> B -> C -> D -> E -> F -> G -> H -> I -> J}
\end{center}

A more visually effective representation of the matrix is presented below, for the understanding of the users, although the above plain text version is used for prompting the LLMs. The numbers in red depict the connections that constitute the solution to the corresponding graph traversal. 



\begin{center}
\begin{tabular}{c|cccccccccc}
& \textbf{~A~} & \textbf{~B~} & \textbf{~C~} & \textbf{~D~} & \textbf{~E~} & \textbf{~F~} & \textbf{~G~} & \textbf{~H~} & \textbf{~I~} & \textbf{~J~} \\  \hline
\textbf{~A~} & ~0 & \textbf{\textbf{\textcolor{red}{~1}}} & ~0 & ~0 & ~0 & ~0 & ~0 & ~0 & ~0 & ~0 \\
\textbf{~B~} & ~0 & ~0 & \textbf{\textcolor{red}{~1}} & ~0 & ~0 & ~0 & ~0 & ~0 & ~0 & ~0 \\
\textbf{~C~} & ~0 & ~0 & ~0 & \textbf{\textcolor{red}{~1}} & ~0 & ~0 & ~0 & ~0 & ~0 & ~0 \\
\textbf{~D~} & ~0 & ~0 & ~0 & ~0 & \textbf{\textcolor{red}{~1}} & ~0 & ~0 & ~0 & ~0 & ~0 \\
\textbf{~E~} & ~0 & ~0 & ~0 & ~0 & ~0 & \textbf{\textcolor{red}{~1}} & ~0 & ~0 & ~0 & ~0 \\
\textbf{~F~} & ~0 & ~0 & ~0 & ~0 & ~0 & ~0 & \textbf{\textcolor{red}{~1}} & ~0 & ~0 & ~0 \\
\textbf{~G~} & ~0 & ~0 & ~0 & ~0 & ~0 & ~0 & ~0 & \textbf{\textcolor{red}{~1}} & ~0 & ~0 \\
\textbf{~H~} & ~0 & ~0 & ~0 & ~0 & ~0 & ~0 & ~0 & ~0 & \textbf{\textcolor{red}{~1}} & ~0 \\
\textbf{~I~} & ~0 & ~0 & ~0 & ~0 & ~0 & ~0 & ~0 & ~0 & ~0 & \textbf{\textcolor{red}{~1}} \\
\textbf{~J~} & ~0 & ~0 & ~0 & ~0 & ~0 & ~0 & ~0 & ~0 & ~0 & ~0 \\
\end{tabular}
\end{center}

\subsubsection{Adding k-shot prompts}
Particularly for k-shot prompts (1-shot and 3-shot settings), we prepend randomly generated prompts to the task, as below. 

\textit{Given is the adjacency matrix for a unweighted undirected graph containing 10 nodes labelled A to J. The value corresponding to each row M and column N represents whether there is a connection between the two nodes, where 0 means no connection.   }

\textit{Consider some examples}

\textit{Example 1: What is the shortest path from node A to node O? Return the sequence of nodes in response.}




\begin{center}
     \begin{tabular}{cccccccccccccccc}

& ~A & ~B & ~C & ~D & ~E & ~F & ~G & ~H & ~I & ~J & ~K & ~L & ~M & ~N & ~O \\
~A & ~0 & ~1 & ~0 & ~1 & ~0 & ~0 & ~0 & ~0 & ~0 & ~0 & ~0 & ~0 & ~1 & ~0 & ~0 \\
~B & ~1 & ~0 & ~1 & ~0 & ~0 & ~0 & ~0 & ~0 & ~0 & ~0 & ~0 & ~0 & ~0 & ~0 & ~0 \\
~C & ~0 & ~1 & ~0 & ~0 & ~0 & ~0 & ~0 & ~0 & ~0 & ~0 & ~0 & ~0 & ~0 & ~0 & ~0 \\
~D & ~1 & ~0 & ~0 & ~0 & ~1 & ~0 & ~0 & ~0 & ~1 & ~0 & ~0 & ~1 & ~0 & ~0 & ~0 \\
~E & ~0 & ~0 & ~0 & ~1 & ~0 & ~1 & ~0 & ~0 & ~0 & ~0 & ~0 & ~0 & ~0 & ~0 & ~0 \\
~F & ~0 & ~0 & ~0 & ~0 & ~1 & ~0 & ~1 & ~0 & ~0 & ~0 & ~0 & ~0 & ~0 & ~0 & ~0 \\
~G & ~0 & ~0 & ~0 & ~0 & ~0 & ~1 & ~0 & ~1 & ~0 & ~0 & ~0 & ~0 & ~0 & ~0 & ~0 \\
~H & ~0 & ~0 & ~0 & ~0 & ~0 & ~0 & ~1 & ~0 & ~0 & ~0 & ~0 & ~0 & ~0 & ~0 & ~0 \\
~I & ~0 & ~0 & ~0 & ~1 & ~0 & ~0 & ~0 & ~0 & ~1 & ~0 & ~0 & ~0 & ~0 & ~0 & ~0 \\
~J & ~0 & ~0 & ~0 & ~0 & ~0 & ~0 & ~0 & ~1 & ~0 & ~1 & ~0 & ~0 & ~0 & ~0 & ~0 \\
~K & ~0 & ~0 & ~0 & ~0 & ~0 & ~0 & ~0 & ~0 & ~1 & ~0 & ~1 & ~0 & ~0 & ~0 & ~0 \\
~L & ~0 & ~0 & ~0 & ~1 & ~0 & ~0 & ~0 & ~0 & ~0 & ~0 & ~0 & ~0 & ~0 & ~0 & ~0 \\
~M & ~1 & ~0 & ~0 & ~0 & ~0 & ~0 & ~0 & ~0 & ~0 & ~0 & ~0 & ~0 & ~1 & ~0 & ~0 \\
~N & ~0 & ~0 & ~0 & ~0 & ~0 & ~0 & ~0 & ~0 & ~0 & ~0 & ~0 & ~1 & ~0 & ~1 & ~0 \\
~O & ~0 & ~0 & ~0 & ~0 & ~0 & ~0 & ~0 & ~0 & ~0 & ~0 & ~0 & ~0 & ~1 & ~0 & ~0 \\

\end{tabular}
\end{center}

\textit{Solution: A -> M -> N -> O} 

\textit{ Given these examples, answer the following quesiton.
}
\textit{What is the shortest path from node A to node J? Return the sequence of nodes in response.}


\begin{center}
\begin{tabular}{ccccccccccc}
 & ~A & ~B & ~C & ~D & ~E & ~F & ~G & ~H & ~I & ~J \\
~A & ~0 & ~1 & ~0 & ~0 & ~0 & ~0 & ~0 & ~0 & ~0 & ~0 \\
~B & ~1 & ~0 & ~1 & ~0 & ~0 & ~1 & ~0 & ~0 & ~0 & ~0 \\
~C & ~0 & ~1 & ~0 & ~1 & ~0 & ~0 & ~0 & ~0 & ~0 & ~0 \\
~D & ~0 & ~0 & ~1 & ~0 & ~1 & ~0 & ~0 & ~0 & ~0 & ~0 \\
~E & ~0 & ~0 & ~0 & ~1 & ~0 & ~0 & ~0 & ~0 & ~0 & ~0 \\
~F & ~0 & ~1 & ~0 & ~0 & ~0 & ~0 & ~1 & ~0 & ~0 & ~0 \\
~G & ~0 & ~0 & ~0 & ~0 & ~0 & ~1 & ~0 & ~1 & ~0 & ~0 \\
~H & ~0 & ~0 & ~0 & ~0 & ~0 & ~0 & ~1 & ~0 & ~1 & ~0 \\
~I & ~0 & ~0 & ~0 & ~0 & ~0 & ~0 & ~0 & ~1 & ~0 & ~1 \\
~J & ~0 & ~0 & ~0 & ~0 & ~0 & ~0 & ~0 & ~0 & ~1 & ~0 \\
\end{tabular}
\end{center}

\subsubsection{Node Jumbling}
An important variant of our graph traversal tasks involves jumbling the labels of the node order. This is primarily done, in order to avoid pretraining distribution biases in the model responses. For example, consider the adjacency matrix



\begin{center}
     \begin{tabular}{ccccccccccc}
& ~A& ~B & ~C &~D &~E &~F& ~G &~H &~I &~J\\
~A & ~0& ~1& ~0& ~0& ~0& ~0& ~0& ~0& ~0& ~0 \\
~B& ~1& ~0& ~1& ~0 &~0 &~0 &~0 &~0 &~0 &~0 \\

~C& ~0& ~1& ~0& ~1& ~0& ~0& ~0& ~0& ~0& ~0 \\
~D& ~0& ~0& ~1& ~0& ~1& ~0 &~0 &~0& ~0& ~0 \\
 ~E &~0 &~0 &~0 & ~1& ~0& ~1 &~0& ~0& ~0& ~0 \\
 ~F& ~0 &~0 &~0 &~0 &~1& ~0& ~1& ~0& ~0 &~0 \\
 ~G &~0 &~0& ~0& ~0& ~0& ~1& ~0& ~1& ~0& ~0 \\
 ~H &~0& ~0& ~0& ~0& ~0& ~0& ~1& ~0 &~1& ~0  \\
 ~I &~0& ~0& ~0& ~0& ~0& ~0& ~0& ~1& ~0& ~1 \\
 ~J& ~0& ~0& ~0 &~0 &~0 &~0& ~0& ~0 &~1 &~0 \\
\end{tabular}
\end{center}

On jumbling, the new matrix looks something like 



\begin{center}
     \begin{tabular}{ccccccccccc}
& ~H& ~C & ~D &~F &~I &~E& ~G &~A &~J &~B\\
~H & ~0& ~1& ~0& ~0& ~0& ~0& ~0& ~0& ~0& ~0 \\
~C& ~1& ~0& ~1& ~0 &~0 &~0 &~0 &~0 &~0 &~0 \\

~D& ~0& ~1& ~0& ~1& ~0& ~0& ~0& ~0& ~0& ~0 \\
~F& ~0& ~0& ~1& ~0& ~1& ~0 &~0 &~0& ~0& ~0 \\
 ~I &~0 &~0 &~0 & ~1& ~0& ~1 &~0& ~0& ~0& ~0 \\
 ~E& ~0 &~0 &~0 &~0 &~1& ~0& ~1& ~0& ~0 &~0 \\
 ~G &~0 &~0& ~0& ~0& ~0& ~1& ~0& ~1& ~0& ~0 \\
 ~A &~0& ~0& ~0& ~0& ~0& ~0& ~1& ~0 &~1& ~0  \\
 ~J &~0& ~0& ~0& ~0& ~0& ~0& ~0& ~1& ~0& ~1 \\
 ~B& ~0& ~0& ~0 &~0 &~0 &~0& ~0& ~0 &~1 &~0 \\
\end{tabular}
\end{center}

which is essentially a renaming of node labels. Correspondingly, the modified task now involves finding the shortest node between the new "first" node and the new "last" node. Unsurprisingly, we observe a drop in the performance of models when we introduce this variation, indicating at least some effect of pretraining data distribution bias (in the form of sequential nodel labeling \textit{A,B,C,...}) in graph traversal tasks.

\subsubsection{Weighted and Directed graph representations}
While weighted connections are represented by integers $\geq 1$, directed connections are represented by removing symmetricity accross the diagonal of the matrix. A simple example to illustrate this is shown below.

\textit{Given is the adjacency matrix for a weighted directed graph containing 16 nodes labelled A to P. The value corresponding to each row M and column N represents the cost of travelling between the two nodes, where 0 means no connection.   
}

\textit{What is the least cost path from node A to node P? Return the sequence of nodes in response.
}



\begin{center}
     \begin{tabular}{ccccccccccccccccc}
 & ~A & ~B & ~C & ~D & ~E & ~F & ~G & ~H & ~I & ~J & ~K & ~L & ~M & ~N & ~O & ~P \\
~A & ~0 & ~3 & ~0 & ~0 & ~0 & ~0 & ~0 & ~0 & ~0 & ~0 & ~0 & ~0 & ~0 & ~0 & ~0 & ~0 \\
~B & ~0 & ~0 & ~4 & ~0 & ~0 & ~0 & ~0 & ~0 & ~0 & ~0 & ~0 & ~0 & ~0 & ~0 & ~0 & ~0 \\
~C & ~0 & ~0 & ~0 & ~2 & ~0 & ~0 & ~0 & ~0 & ~0 & ~0 & ~0 & ~0 & ~0 & ~0 & ~0 & ~0 \\
~D & ~0 & ~0 & ~0 & ~0 & ~0 & ~0 & ~0 & ~2 & ~0 & ~0 & ~0 & ~0 & ~0 & ~0 & ~0 & ~0 \\
~E & ~2 & ~0 & ~0 & ~0 & ~0 & ~0 & ~0 & ~0 & ~1 & ~0 & ~0 & ~0 & ~0 & ~0 & ~0 & ~0 \\
~F & ~0 & ~0 & ~0 & ~0 & ~1 & ~0 & ~0 & ~0 & ~0 & ~4 & ~0 & ~0 & ~0 & ~0 & ~0 & ~0 \\
~G & ~0 & ~0 & ~3 & ~0 & ~0 & ~4 & ~0 & ~0 & ~0 & ~0 & ~0 & ~0 & ~0 & ~0 & ~0 & ~0 \\
~H & ~0 & ~0 & ~0 & ~0 & ~0 & ~0 & ~0 & ~0 & ~0 & ~0 & ~0 & ~1 & ~0 & ~0 & ~0 & ~0 \\
~I & ~0 & ~0 & ~0 & ~0 & ~0 & ~0 & ~0 & ~0 & ~0 & ~5 & ~0 & ~0 & ~2 & ~0 & ~0 & ~0 \\
~J & ~0 & ~0 & ~0 & ~0 & ~0 & ~0 & ~0 & ~0 & ~0 & ~0 & ~0 & ~0 & ~0 & ~2 & ~0 & ~0 \\
~K & ~0 & ~0 & ~0 & ~0 & ~0 & ~0 & ~1 & ~0 & ~0 & ~5 & ~0 & ~0 & ~0 & ~0 & ~0 & ~0 \\
~L & ~0 & ~0 & ~0 & ~0 & ~0 & ~0 & ~0 & ~0 & ~0 & ~0 & ~3 & ~0 & ~0 & ~0 & ~0 & ~0 \\
~M & ~0 & ~0 & ~0 & ~0 & ~0 & ~0 & ~0 & ~0 & ~0 & ~0 & ~0 & ~0 & ~0 & ~2 & ~0 & ~0 \\
~N & ~0 & ~0 & ~0 & ~0 & ~0 & ~0 & ~0 & ~0 & ~0 & ~0 & ~0 & ~0 & ~0 & ~0 & ~2 & ~0 \\
~O & ~0 & ~0 & ~0 & ~0 & ~0 & ~0 & ~0 & ~0 & ~0 & ~0 & ~2 & ~0 & ~0 & ~0 & ~0 & ~2 \\
~P & ~0 & ~0 & ~0 & ~0 & ~0 & ~0 & ~0 & ~0 & ~0 & ~0 & ~0 & ~0 & ~0 & ~0 & ~0 & ~0 \\

\end{tabular}
\end{center}

\begin{table}[]
\caption{Evaluation of models on problem category 1: tree-based graph traversals. The values are represented as A/B, where A depicts the average normalized binary accuracy (ranging from 0 to 1) for 10 examples per setting, and B depicts the average normalized value for partial credit for 10 examples per setting. \textbf{*} = a larger context variant of the corresponding LLM was used. \textbf{$\bigtriangleup$} = the prompt was too large to fit the context window of the model, hence the setting was omitted. \textbf{**} = the model did not return any solution and refused to answer the question.} 
\label{tab:prob1}
\centering
\resizebox{\linewidth}{!}{
\begin{tabular}{llllllllll}
\hline
\textbf{} & \multicolumn{3}{l}{\textbf{O(10)}} & \multicolumn{3}{l}{\textbf{O(20)}} & \multicolumn{3}{l}{\textbf{O(20) jumbled}} \\ \hline
Problem & \textbf{0-shot} & \textbf{1-shot} & \textbf{3-shot} & \textbf{0-shot} & \textbf{1-shot} & \textbf{3-shot} & \textbf{0-shot} & \textbf{1-shot} & \textbf{3-shot} \\ \hline
\textbf{} & \multicolumn{9}{c}{\textbf{GPT-3.5}} \\ \hline
\textbf{1.1} & 1.0 & 0.6/0.80 & 1.0 & 1.0 & 1.0 & 1.0* & 0.0/0.09 & 0.1/0.08 & 0.0/0.07* \\
\textbf{1.2} & 0.2/0.41 & 0.4/0.59 & 0.6/0.62 & 0.0/0.39 & 0.1/0.38 & 0.2/0.31* & 0.0/0.22 & 0.2/0.29 & 0.0/0.29* \\
\textbf{1.3} & 0.3/0.53 & 0.0/0.24 & 0.4/0.48 & 0.2/0.36 & 0.1/0.17 & 0.0/0.26* & 0.0/0.28 & 0.0/0.21 & 0.0/0.19* \\
\textbf{1.4} & 0.3/0.47 & 0.2/0.45 & 0.0/0.48 & 0.1/0.37 & 0.0/0.23 & 0.0/0.33* & 0.1/0.24 & 0.0/0.13 & 0.0/29* \\
\hline
\textbf{} & \multicolumn{9}{c}{\textbf{GPT-4}} \\ \hline
\textbf{1.1} & 1.0 & 1.0 & 1.0 & 1.0 & 1.0 & 1.0 & 0.7/0.84 & 0.9/0.98 & 0.7/0.87 \\
\textbf{1.2} & 1.0 & 1.0 & 1.0 & 0.5/0.67 & 0.8/0.85 & 0.9/0.93 & 0.1/0.31 & 0.2/0.47 & 0.4/0.55 \\
\textbf{1.3} & 0.8/0.93 & 0.6/0.69 & 0.9/0.91 & 0.4/0.64 & 0.5/0.70 & 0.6/0.71 & 0.2/0.45 & 0.2/0.40 & 0.3/0.51 \\
\textbf{1.4} & 0.4/0.62 & 0.0/0.31 & 0.4/0.59 & 0.2/0.40 & 0.1/0.35 & 0.0/0.27 & 0.1/0.26 & 0.2/0.39 & 0.3/0.51 \\
\hline
\textbf{} & \multicolumn{9}{c}{\textbf{Claude-2}} \\ \hline
\textbf{1.1} & 1.0 & 0.0/0.60 & 1.0 & 0.2/0.30 & 0.2/0.43 & 1.0 & 0.0/0.07 & 0.0/0.06 & 0.0/0.12 \\
\textbf{1.2} & 0.7/0.84 & 0.0/0.30 & 0.3/0.50 & 0.3/0.53 & 0.1/0.29 & 0.2/0.31 & 0.1/0.31 & 0.0/0.18 & 0.0/0.22 \\
\textbf{1.3} & 0.4/0.62 & 0.5/0.70 & 0.2/0.56 & 0.0/0.37 & 0.2/0.39 & 0.1/0.35 & 0.3/0.36 & 0.0/0.26 & 0.0/0.14 \\
\textbf{1.4} & 0.4/0.56 & 0.2/0.44 & 0.1/0.34 & 0.2/0.41 & 0.0/0.25 & 0.1/0.34 & 0.1/0.30 & 0.0/0.19 & 0.0/0.17 \\
\hline
\textbf{} & \multicolumn{9}{c}{\textbf{Llama-2}} \\ \hline
\textbf{1.1} & 1.0 & 0.0/0.10 & 1.0 & 0.0/0.05 & 0.0/1.0 &  $\bigtriangleup$ & 0.0/0.06 & 0.0/0.05 & $\bigtriangleup$ \\
\textbf{1.2} & 0.2/0.47 & 0.1/0.31 & 0.0/0.18 & 0.0/0.15 & 0.0/0.14 & $\bigtriangleup$ & 0.0/0.11 & 0.0/0.16 & $\bigtriangleup$ \\
\textbf{1.3} & 0.0/0.30 & 0.1/0.35 & 0.0/0.25 & 0.0/0.17 & 0.0/0.13 & $\bigtriangleup$ & 0.0/0.14 & 0.0/0.16 & $\bigtriangleup$ \\
\textbf{1.4} & 0.2/0.40 & 0.1/0.37 & 0.0/0.32 & 0.0/0.16 & 0.0/0.25 & $\bigtriangleup$ & 0.0/0.15 & 0.0/0.17 & $\bigtriangleup$ \\
\hline
\textbf{} & \multicolumn{9}{c}{\textbf{Palm-2}} \\ \hline
\textbf{1.1} & 1.0 & 1.0 & 1.0 & 1.0 & ** & ** & 0.0/0.05 & 0.0/0.05 & 0.0/0.06 \\
\textbf{1.2} & 0.2/0.27 & 0.0/0.32 & 0.1/0.44 & 0.0/0.27 & ** & ** & 0.0/0.19 & 0.0/0.17 & ** \\
\textbf{1.3} & 0.1/0.52 & 0.2/0.53 & 0.2/0.50 & ** & ** & ** & ** & ** & ** \\
\textbf{1.4} & 0.0/0.35 & 0.1/0.38 & 0.0/0.34 & 0.0/0.10 & ** & ** & 0.0/0.08 & ** & **
\end{tabular}
}
\end{table}

\begin{table}[]
\caption{Evaluation of models on problem category 2: grid-based graph traversals.  The values are represented as A/B, where A depicts the average normalized binary accuracy (ranging from 0 to 1) for 10 examples per setting, and B depicts the average normalized value for partial credit for 10 examples per setting. The notations are similar to Table \ref{tab:prob1}.}
\label{tab:prob2}
\centering
\resizebox{\linewidth}{!}{
\begin{tabular}{llllllllll}
\hline
\textbf{} & \multicolumn{3}{l}{\textbf{O(10)}} & \multicolumn{3}{l}{\textbf{O(20)}} & \multicolumn{3}{l}{\textbf{O(20) jumbled}} \\ \hline
\textbf{Problem} & \textbf{0-shot} & \textbf{1-shot} & \textbf{3-shot} & \textbf{0-shot} & \textbf{1-shot} & \textbf{3-shot} & \textbf{0-shot} & \textbf{1-shot} & \textbf{3-shot} \\ \hline
\textbf{} & \multicolumn{9}{c}{\textbf{GPT-3.5}} \\ \hline
\textbf{2.1} & 0.1/0.33 & 0.2/0.46 & 0.3/0.41 & 0.1/0.21 & 0.1/0.28 & 0.1/0.28* & 0.1/0.14 & 0.0/0.13 & 0.0/0.12* \\
\textbf{2.2} & 0.0/0.40 & 0.0/0.19 & 0.0/0.37 & 0.0/0.30 & 0.0/0.12 & 0.1/0.32* & 0.0/0.13 & 0.1/0.13 & 0.0/0.13* \\
\textbf{2.3} & 0.1/0.40 & 0.10/0.36 & 0.0/0.37 & 0.0/0.14 & 0.0/0.17 & 0.0/0.22* & 0.0/0.13 & 0.0/0.09 & 0.0/0.11* \\
\textbf{2.4} & 0.0 & 0.0 & 0.0 & 0.0 & 0.0 & 0.3* & 0.0 & 0.0 & 0.0* \\
\hline
\textbf{} & \multicolumn{9}{c}{\textbf{GPT-4}} \\ \hline
\textbf{2.1} & 0.3/0.46 & 0.3/0.48 & 0.4/0.51 & 0.1/0.36 & 0.1/0.29 & 0.0/0.21 & 0.0/0.14 & 0.0/0.23 & 0.0/0.16 \\
\textbf{2.2} & 0.0/0.44 & 0.4/0.56 & 0.2/0.37 & 0.1/0.42 & 0.1/0.31 & 0.3/0.54 & 0.0/0.14 & 0.0/0.26 & 0.0/0.23 \\
\textbf{2.3} & 0.2/0.52 & 0.0/0.36 & 0.2/0.40 & 0.1/0.34 & 0.2/0.49 & 0.1/0.28 & 0.0/0.16 & 0.0/0.16 & 0.0/0.17 \\
\textbf{2.4} & 0.3 & 0.3 & 0.4 & 0.4 & 0.4 & 0.4 & 0.4 & 0.9 & 0.6 \\
\hline
\textbf{} & \multicolumn{9}{c}{\textbf{Claude-2}} \\ \hline
\textbf{2.1} & 0.0/0.40 & 0.3/0.44 & 0.0/0.36 & 0.0/0.29 & 0.1/0.20 & 0.0/0.22 & 0.0/0.16 & 0.0/0.13 & 0.1/0.16 \\
\textbf{2.2} & 0.1/0.34 & 0.2/0.41 & 0.1/0.31 & 0.0/0.19 & 0.3/0.23 & 0.0/0.21 & 0.0/0.17 & 0.1/0.12 & 0.0/0.13 \\
\textbf{2.3} & 0.1/0.33 & 0.1/0.29 & 0.0/0.24 & 0.0/0.17 & 0.0/0.18 & 0.0/0.26 & 0.0/0.12 & 0.0/0.13 & 0.0/0.12 \\
\textbf{2.4} & 0.0 & 0.1 & 0.3 & 0.0 & 0.1 & 0.3 & 0.0 & 0.3 & 0.4 \\
 \hline
\textbf{} & \multicolumn{9}{c}{\textbf{Llama-2}} \\ \hline
\textbf{2.1} & 0.0/0.16 & 0.2/0.44 & 0.1.0.24 & 0.0/0.18 & 0.1/0.12 & $\bigtriangleup$ & 0.0/0.05 & 0.0/0.05 & $\bigtriangleup$ \\
\textbf{2.2} & 0.0/0.22 & 0.0/0.40 & 0.1/0.52 & 0.0/0.18 & 0.0/0.36 & $\bigtriangleup$ & 0.0/0.04 & 0.0/0.06 & $\bigtriangleup$ \\
\textbf{2.3} & 0.0/0.14 & 0.0/0.22 & 0.0/0.12 & 0.0/0.08 & 0.0/0.12 & $\bigtriangleup$ & 0.0/0.014 & 0.0/0.06 & $\bigtriangleup$ \\
\textbf{2.4} & 0.1 & 0.8 & 0.0 & 0.0 & 0.0 & $\bigtriangleup$ & 0.0 & 0.0 & $\bigtriangleup$ \\
 \hline
\textbf{} & \multicolumn{9}{c}{\textbf{Palm-2}} \\ \hline
\textbf{2.1} & 0.0/0.22 & 0.5/0.64 & 0.2/0.42 & 0.0/0.06 & ** & ** & 0.0/0.04 & ** & ** \\
\textbf{2.2} & 0.0/0.21 & 0.0/0.24 & 0.0/0.30 & 0.0/0.04 & ** & ** & ** & ** & ** \\
\textbf{2.3} & 0.0/0.18 & 0.0/0.30 & 0.0/0.22 & ** &**  &**  & 0.0/0.08 & ** & ** \\
\textbf{2.4} & 0.0 & 0.0 & 0.0 & ** & ** & ** & ** & ** & **
\end{tabular}
}
\end{table}

\begin{table}[]
\caption{Evaluation of models on problem category 3: special problems. The values depict the average normalized value (ranging from 0 to 1) for 10 examples per setting. \textbf{-} = the setting is not applicable for a particular level. \textbf{$\bigtriangleup$} = the prompt was too large to fit the context window of the model, hence the setting was ommitted. \textbf{**} = the model did not return any solution and refused to answer the question. Note that for problem 3.1, the random baseline is 0.5, since the problem solution is binary (True/False).}
\label{tab:prob3}
\centering
\resizebox{0.7\linewidth}{!}{%
\begin{tabular}{llllllllll}
\hline
 & \multicolumn{3}{l}{\textbf{O(10)}} & \multicolumn{3}{l}{\textbf{O(20)}} & \multicolumn{3}{l}{\textbf{O(20) jumbled}} \\ \hline
\textbf{Problem} & \textbf{0-shot} & \textbf{1-shot} & \textbf{3-shot} & \textbf{0-shot} & \textbf{1-shot} & \textbf{3-shot} & \textbf{0-shot} & \textbf{1-shot} & \textbf{3-shot} \\ \hline
 & \multicolumn{9}{c}{\textbf{GPT-3.5}} \\ \hline
\textbf{3.1} & 0.3 & 0.1 & 0.5 & - & - & - & - & - & - \\
\textbf{3.2} & 0.0 & 0.0 & 0.2 & 0.0 & 0.0 & 0.0 & 0.0 & 0.0 & 0.0 \\ \hline
 & \multicolumn{9}{c}{\textbf{GPT-4}} \\ \hline
\textbf{3.1} & 0.5 & 0.5 & 0.2 & - & - & - & - & - & - \\
\textbf{3.2} & 0.2 & 0.3 & 0.2 & 0.0 & 0.0 & 0.1 & 0.0 & 0.1 & 0.0 \\ \hline
 & \multicolumn{9}{c}{\textbf{Claude-2}} \\ \hline
\textbf{3.1} & 0.6 & 0.4 & 0.7 & - & - & - & - & - & - \\
\textbf{3.2} & 0.1 & 0.1 & 0.2 & 0.1 & 0.0 & 0.0 & 0.0 & 0.0 & 0.0 \\ \hline
 & \multicolumn{9}{c}{\textbf{Llama-2}} \\ \hline
\textbf{3.1} & 0.6 & 0.8 & 0.7 & - & - & - & - & - & - \\
\textbf{3.2} & 0.0 & 0.0 & 0.0 & 0.0 & 0.0 & $\bigtriangleup$  & 0.0 & 0.0 & $\bigtriangleup$  \\ \hline
 & \multicolumn{9}{c}{\textbf{Palm-2}} \\ \hline
\textbf{3.1} & 0.5 & 0.1 & 0.2 & - & - & - & - & - & - \\
\textbf{3.2} & 0.0 & 0.0 & 0.0 & 0.0 & ** & ** & 0.0 & ** & **
\end{tabular}%
}
\end{table}

\begin{table*}[]
\caption{Evaluation of models on O(10) graphs with \textbf{Chain-of-Thought Reasoning}. The values are represented as A/B, where A depicts the average normalized binary accuracy (ranging from 0 to 1) for 10 examples per setting, and B depicts the average normalized value for partial credit for 10 examples per setting. \textbf{*} = a larger context variant of the corresponding LLM was used. \textbf{$\bigtriangleup$} = the prompt was too large to fit the context window of the model, hence the setting was ommitted. \textbf{**} = the model did not return any solution and refused to answer the question.} 
\label{tab:CoT}
\centering
\resizebox{0.9\textwidth}{!}{
\begin{tabular}{llll|llll|llll}
\hline
Prob & \textbf{0-shot} & \textbf{1-shot} & \textbf{3-shot} & Prob & \textbf{0-shot} & \textbf{1-shot} & \textbf{3-shot} & Prob & \textbf{0-shot} & \textbf{1-shot} & \textbf{3-shot} \\ \hline
\textbf{} & \multicolumn{11}{c}{\textbf{GPT-3.5}} \\ \hline
\textbf{1.1} & 0.6/0.65 &  1.0 & 0.9/0.92 & \textbf{2.1} & 0.0/0.20 & 0.2/0.50 & 0.0/0.32*  & \textbf{3.1} & 0.3 & 0.1 & 0.5 \\
\textbf{1.2} & 0.2/0.38& 0.3/0.55 & 0.5/0.72 & \textbf{2.2} & 0.1/0.27 & 0.0/0.37 & 0.1/0.2* & \textbf{3.2}& 0.0&0.0  &0.2  \\
\textbf{1.3} & 0.2/0.42  & 0.12/0.38 & 0.0/0.28 & \textbf{2.3} & 0.0/0.2 & 0.1/0.29 & 0.0/0.18* & & &  &  \\
\textbf{1.4} & 0.0/0.29 & 0.0/0.28  &0.0/0.25 & \textbf{2.4} & 0.0/0.0 & 0.0/0.0  & 0.1/0.1* & & & &  \\
\hline
\textbf{} & \multicolumn{11}{c}{\textbf{GPT-4}} \\ \hline
\textbf{1.1} & 0.7/0.74  & 1.0  & 1.0 & \textbf{2.1} & 0.4/0.62 & 0.3/0.42 & 0.0/0.26 & \textbf{3.1} &0.5 &0.5  & 0.2 \\
\textbf{1.2} & 1.0 & 0.9/0.96 & 0.8/0.88 & \textbf{2.2} & 0.2/0.64 & 0.3/0.63 & 0.4/0.51 & \textbf{3.2} & 0.2& 0.3 & 0.2 \\
\textbf{1.3} & 0.8/0.88 & 0.9/0.91 & 0.9/0.94 & \textbf{2.3} & 0.0/0.34  &0.1/0.45 & 0.6/0.76 &  &  &  &  \\
\textbf{1.4} & 0.6/0.77 & 0.3/0.49 & 0.6/0.74 & \textbf{2.4} & 0.6/0.6 & 0.8/0.8 & 0.7/0.7 & &  &  &  \\
\hline
\textbf{} & \multicolumn{11}{c}{\textbf{Claude-2}} \\ \hline
\textbf{1.1} & 1.0 & 1.0  & 1.0  & \textbf{2.1} & 0.0/0.36 & 0.2/0.49  &0.0/0.29  & \textbf{3.1} & 0.6 & 0.4  &0.7  \\
\textbf{1.2} & 0.9/0.95 & 0.3/0.48 & 0.30/47 & \textbf{2.2} & 0.0/0.34  & 0.0/0.27 & 0.1/0.27 & \textbf{3.2} & 0.1& 0.1 & 0.2 \\
\textbf{1.3} &  0.9/0.92 & 0.5/0.75 & 0.4/0.44 & \textbf{2.3} & 0.0/0.3 & 0.1/0.29 & 0.2/0.39 &  &  & &  \\
\textbf{1.4} & 0.4/0.50& 0.0/0.23 & 0.0/0.28  & \textbf{2.4} & 0.0/0.0 & 0.0/0.0 & 0.1/0.1 & &  &  &  \\
\end{tabular}
}
\end{table*}

\begin{table*}[]
\caption{Evaluation of models on O(10) graphs using the \textbf{PathCompare} method. The values are represented as A/B, where A depicts the average normalized binary accuracy (ranging from 0 to 1) for 10 examples per setting, and B depicts the average normalized value for partial credit for 10 examples per setting. \textbf{*} = a larger context variant of the corresponding LLM was used. \textbf{$\bigtriangleup$} = the prompt was too large to fit the context window of the model, hence the setting was omitted. \textbf{**} = The model did not return any solution and refused to answer the question.} 
\label{tab:pathcompare}
\resizebox{0.9\textwidth}{!}{
\begin{tabular}{llll|llll|llll}
\hline

Prob & \textbf{0-shot} & \textbf{1-shot} & \textbf{3-shot} & Prob & \textbf{0-shot} & \textbf{1-shot} & \textbf{3-shot} & Prob & \textbf{0-shot} & \textbf{1-shot} & \textbf{3-shot} \\ \hline
\textbf{} & \multicolumn{11}{c}{\textbf{GPT-3.5}} \\ \hline
\textbf{1.1} & 0.0/0.02 &  0.0/0.02 & 0.0/0.08 & \textbf{2.1} & 0.0/0.45 & 0.5/0.71 & 0.3/0.57*  & \textbf{3.1} & - & - & - \\
\textbf{1.2} & 0.2/0.51 & 0.4/0.62 & 0.5/0.80 & \textbf{2.2} & 0.0/0.46 & 0.2/0.42 & 0.0/0.31* & \textbf{3.2} & 0.0& 0.2 & 0.3 \\
\textbf{1.3} &  0.4/0.61 & 0.3/0.49 & 0.3/0.66 & \textbf{2.3} & 0.0/0.44 & 0.2/0.46  & 0.3/0.54 &  &  \\
\textbf{1.4} & 0.4/0.52 & 0.3/0.6  & 0.2/0.48  & \textbf{2.4} & 0.0/0.0 & 0.0/0.0 & 0.0/0.0 &  &  &  \\
\hline
\textbf{} & \multicolumn{11}{c}{\textbf{GPT-4}} \\ \hline
\textbf{1.1} & 1.0  & 1.0  & 1.0 & \textbf{2.1} & 0.4/0.69 & 0.4/0.52 &  0.2/0.4 & \textbf{3.1} & -  & - & - \\
\textbf{1.2} & 1.0 & 1.0 & 1.0 & \textbf{2.2} & 0.6/0.7 & 0.2/0.44 & 0.1/0.46 & \textbf{3.2} & 0.2 & 0.2 & 0.3 \\
\textbf{1.3} & 1.0 & 1.0 & 1.0 & \textbf{2.3} & 0.5/0.68 & 0.2/0.54 & 0.2/0.59 &  &  \\
\textbf{1.4} & 0.8/0.89 & 0.8/0.81 & 0.4/0.52 & \textbf{2.4} & 0.3/0.3 & 0.3/0.3 & 0.4/0.4 & &  &  &  \\
\hline
\textbf{} & \multicolumn{11}{c}{\textbf{Claude-2}} \\ \hline
\textbf{1.1} & 1.0 & 1.0 & 1.0  &  \textbf{2.1} & 0.1/0.44 & 0.1/0.52  & 0.0/0.24 &\textbf{3.1}& -  & - & - \\
\textbf{1.2} & 1.0 & 0.1/0.32 & 0.4/0.57 & \textbf{2.2} & 0.1/0.32 & 0.1/0.28 & 0.1/0.34 & \textbf{3.2}  & 0.2 & 0.0 & 0.1 \\
\textbf{1.3} & 0.6/0.69  & 0.3/0.42 & 0.4/0.57 & \textbf{2.3} & 0.2/0.38  & 0.1/0.42& 0.3/38 &  &  & &  \\
\textbf{1.4} & 0.5/0.65 & 0.7/0.85 & 0.3/0.46  & \textbf{2.4} &0.0/0.0& 0.0/0.0 &0.0/0.0 & & &  &  \\


\end{tabular}
}
\end{table*}